\documentclass[10pt,twocolumn,letterpaper]{article}

\usepackage[pagenumbers]{configs/cvpr} 

\usepackage{amsmath,amsfonts,bm}

\def\eqref#1{equation~\ref{#1}}

\def\1{\bm{1}}

\DeclareMathAlphabet{\mathsfit}{\encodingdefault}{\sfdefault}{m}{sl}
\SetMathAlphabet{\mathsfit}{bold}{\encodingdefault}{\sfdefault}{bx}{n}

\usepackage{hyperref}
\usepackage{url}
\usepackage{graphicx}
\usepackage{wrapfig}
\usepackage{tabularx}
\usepackage{color, colortbl}
\usepackage{physics}
\usepackage{booktabs}
\usepackage{multirow}
\usepackage{xcolor}
\usepackage{graphicx}
\usepackage{footnote}
\usepackage{tabularx}
\usepackage{float}
\usepackage{lipsum}
\usepackage{multicol}

\newcommand{\name} {MAGE}
	
\definecolor{LightCyan}{rgb}{0.88,1,1}
\definecolor{Grey}{rgb}{0.93,0.93,0.93}
\definecolor{DarkGrey}{rgb}{0.55,0.55,0.55}

\def\cvprPaperID{2102} 
\def\confName{CVPR}
\def\confYear{2023}

\newcommand{\tianhong}[1]{\textcolor{blue}{\ignorespaces}}
\newcommand{\huiwen}[1]{\textcolor{red}{\ignorespaces}}
\newcommand{\han}[1]{\textcolor{cyan}{\ignorespaces}}
\newcommand{\dilip}[1]{\textcolor{orange}{\ignorespaces}}
\newcommand{\shlok}[1]{\textcolor{green}{\ignorespaces}}

\newcommand{\red}[1]{\textcolor{black}{ #1}}
\newcommand{\grey}[1]{\textcolor{DarkGrey}{ #1}}

\colorlet{darkgreen}{green!65!black}
\colorlet{darkblue}{blue!75!black}
\colorlet{darkred}{red!80!black}
\definecolor{lightblue}{HTML}{0071bc}
\definecolor{lightgreen}{HTML}{39b54a}
\definecolor{shadecolor}{RGB}{150,150,150}
\newcommand{\magecolor}[1]{\par\noindent\colorbox{shadecolor}}

\newenvironment{Itemize}%
{
\setlength{\leftmargini}{9pt}%
\begin{itemize}%
\setlength{\itemsep}{0pt}%
\setlength{\topsep}{0pt}%
\setlength{\partopsep}{0pt}%
\setlength{\parskip}{0pt}}%
{\end{itemize}}

\graphicspath{{./figures/}}

\begin{document}

\title{MAGE: MAsked Generative Encoder to Unify \\
Representation Learning and Image Synthesis}

\author{Tianhong Li$^1$\thanks{This work was done when Tianhong Li was intern at Google Research. Correspondence to: Tianhong Li \texttt{<tianhong@mit.edu>}, Huiwen Chang \texttt{<huiwenchang@google.com>}.}
\quad Huiwen Chang$^3$
\quad Shlok Kumar Mishra$^2$
\quad Han Zhang$^3$ \\
Dina Katabi$^1$
\quad  Dilip Krishnan$^3$ \\
$^1$MIT CSAIL \quad $^2$University of Maryland \quad $^3$Google Research
}

\maketitle

\begin{abstract}
Generative modeling and representation learning are two key tasks in computer vision. 
However, these models are typically trained independently, which ignores the potential for each task to help the other, and leads to training and model maintenance overheads.
In this work, we propose MAsked Generative Encoder (\name), the first framework to unify SOTA image generation and self-supervised representation learning.  Our key insight is that using variable masking ratios in masked image modeling pre-training can allow generative training (very high masking ratio) and representation learning (lower masking ratio) under the same training framework. Inspired by previous generative models, MAGE uses semantic tokens learned by a vector-quantized GAN at inputs and outputs, combining this with masking. 
We can further improve the representation by adding a contrastive loss to the encoder output. We extensively evaluate the generation and representation learning capabilities of \name. On ImageNet-1K, \red{a single \name~ViT-L model} obtains \textbf{9.10} FID in the task of class-unconditional image generation and \textbf{78.9\%} top-1 accuracy for linear probing, achieving state-of-the-art performance in both image generation and representation learning. Code is available at \href{https://github.com/LTH14/mage}{\textcolor{red}{\texttt{https://github.com/LTH14/mage}}}.

\end{abstract}

\section{Introduction}

In recent years, we have seen rapid progress in both generative models and representation learning of visual data. Generative models have demonstrated increasingly spectacular performance in generating realistic images \cite{chang2022maskgit, brock2018large, dhariwal2021diffusion, rombach2022high}, while state-of-the-art self-supervised representation learning methods can extract representations at a high semantic level to achieve excellent performance on a number of downstream tasks such as linear probing and few-shot transfer \cite{chen2020simple, chen2021empirical, MAE, bao2021beit, grill2020bootstrap, caron2021emerging}.

\begin{figure}[t]
\begin{center}
\includegraphics[width=0.48\textwidth]{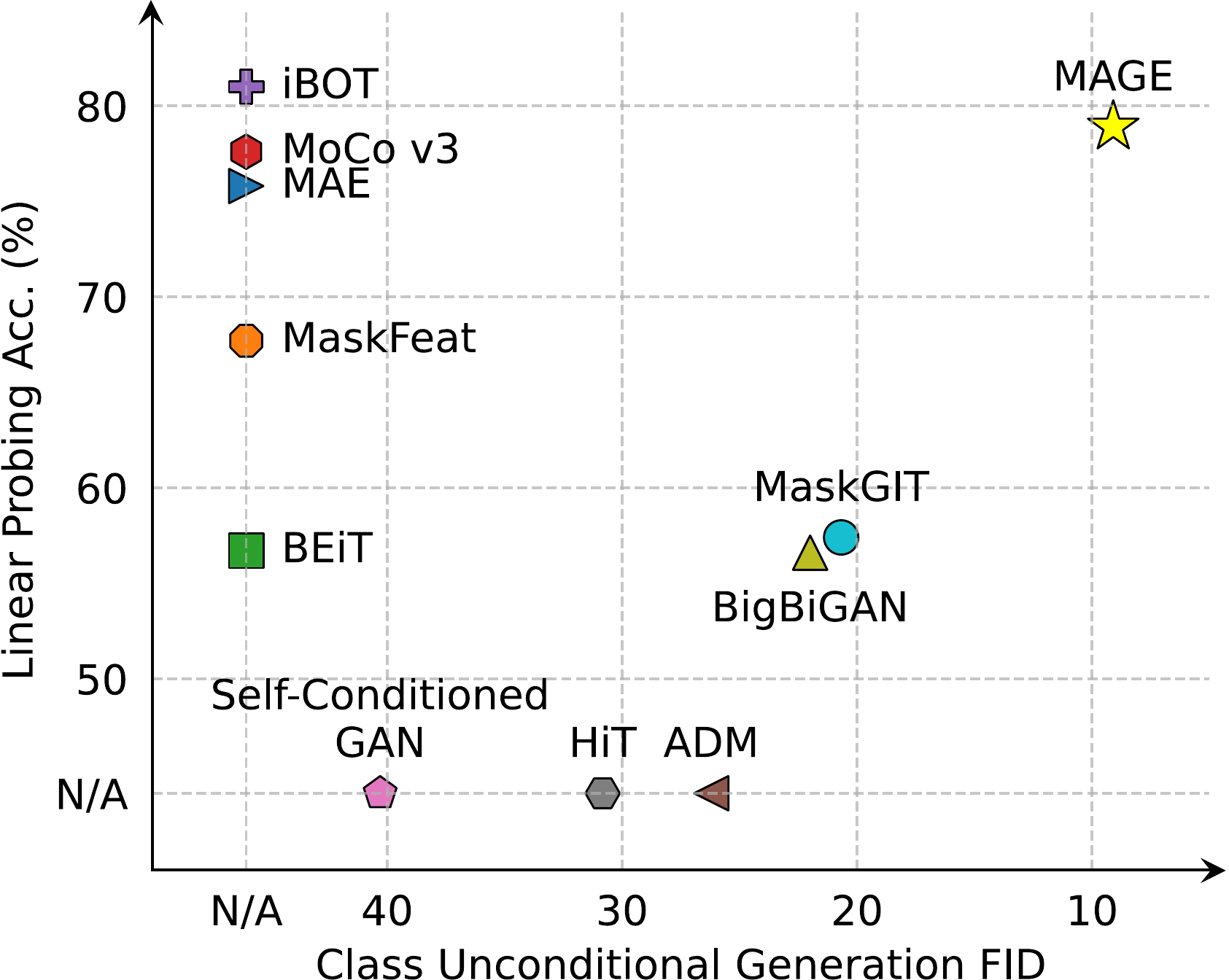}
\end{center}
\vspace{-15pt}
\caption[]{Linear probing and class unconditional generation performance of different methods trained and evaluated on ImageNet-1K. \name~achieves SOTA performance in linear probing and establishes a new SOTA in class unconditional generation.}
\vspace{-15pt}
\label{fig:lp-gen}
\end{figure}

\begin{figure*}[t]
\begin{center}
\vspace{-5pt}
\includegraphics[width=1\textwidth]{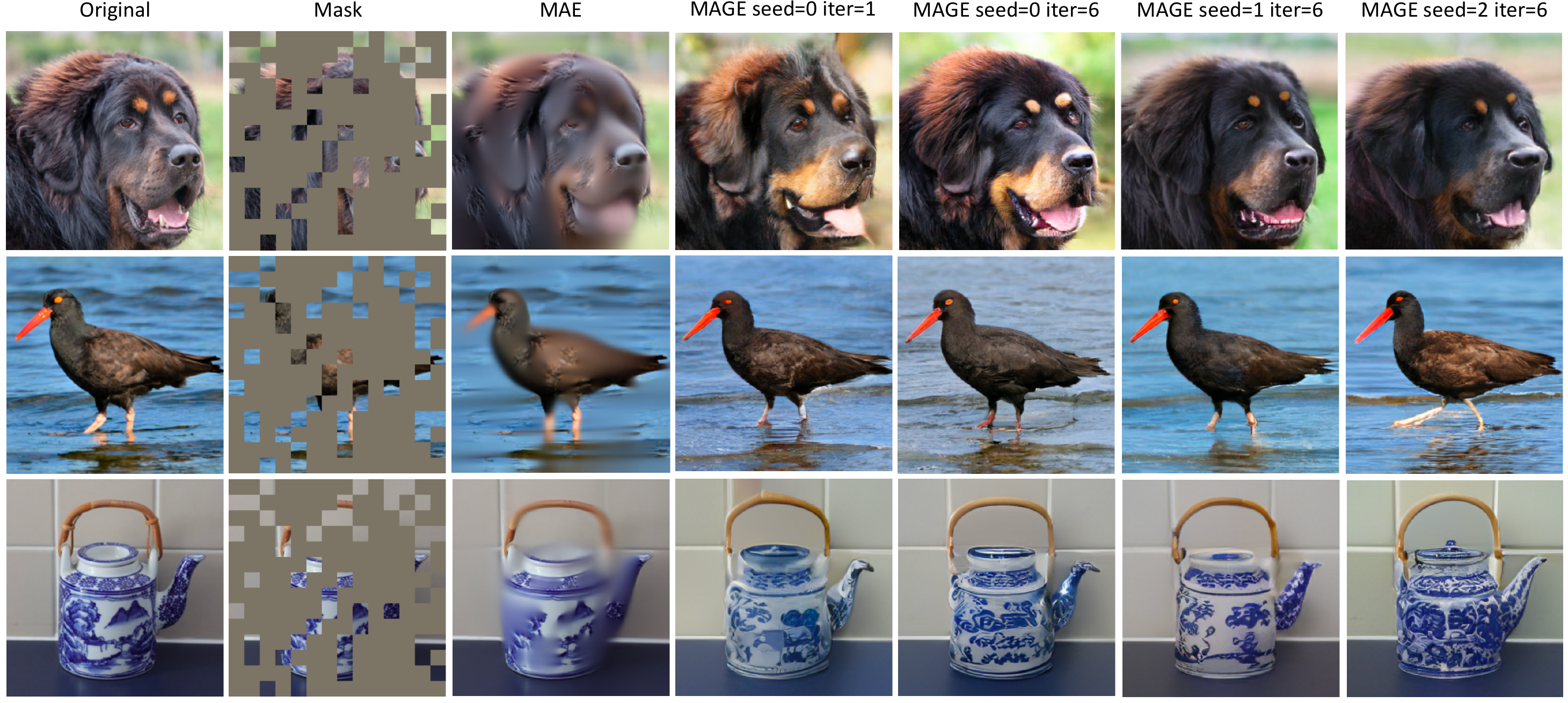}
\end{center}
\vspace{-10pt}
\caption[]{Reconstruction results using MAE and \name~with 75\% masking ratio. MAE reconstructs blurry images with low quality, while \name~can reconstruct high-quality images with detail, and further improves quality through iterative decoding (see \autoref{sec:evaluation} for details). With the same mask, \name~generates diverse reconstruction results with different random seeds. 
Note that the mask for \name~is on semantic tokens whereas that of MAE is on patches in the input image.}
\vspace{-5pt}
\label{fig:teaser}
\end{figure*}

Currently, these two families of models are typically trained independently. 
Intuitively, since generation and recognition tasks require both visual and semantic understanding of data, they should be complementary when combined in a single framework. Generation benefits representation by ensuring that both high-level semantics and low-level visual details are captured; conversely, representation benefits generation by providing rich semantic guidance.  Researchers in natural language processing have observed this synergy: frameworks such as BERT \cite{devlin2018bert} have both high-quality text generation and feature extraction. Another example is DALLE-2 \cite{ramesh2022hierarchical}, where latents conditioned on a \emph{pre-trained} CLIP representation are used to create high-quality text-to-image generations. 

However, in computer vision, there are currently no widely adopted models that unify image generation and representation learning in the same framework.
Such unification is non-trivial due to the structural difference between these tasks: in generative modeling, we \textbf{output} high-dimensional data, conditioned on low-dimension inputs such as class labels, text embeddings, or random noise. In representation learning, we \textbf{input} a high-dimensional image and create a low-dimensional compact embedding useful for downstream tasks. 

Recently, a number of papers have shown that representation learning frameworks based on masked image modeling (MIM) can obtain high-quality representations \cite{MAE, CMAE, bao2021beit, dong2021peco}, often with very high masking ratios (e.g. $75\%$) \cite{MAE}. Inspired by NLP, these methods mask some patches at the input, and the pre-training task is to reconstruct the original image by predicting these masked patches. After pre-training, task-specific heads can be added to the encoder to perform linear probe or fine-tuning. 

These works inspire us to revisit the unification question. Our key insight in this work is that generation is viewed as ``reconstructing" images that are $100\%$ masked, while representation learning is viewed as ``encoding" images that are $0\%$ masked. We can therefore enable a unified architecture by using a \emph{variable masking ratio} during MIM pre-training. The model is trained to reconstruct over a \emph{wide range} of masking ratios covering high masking ratios that enable generation capabilities, and lower masking ratios that enable representation learning. This simple but very effective approach allows a smooth combination of generative training and representation learning in the \emph{same framework}: same architecture, training scheme, and loss function.

However, directly combining existing MIM methods with a variable masking ratio is insufficient for high quality generation because such methods typically use a simple reconstruction loss on pixels, leading to blurry outputs.
For example, as a representative of such methods, the reconstruction quality of MAE \cite{he2022masked} is poor: fine details and textures are missing (\autoref{fig:teaser}). A similar issue exists in many other MIM methods \cite{chen2022context, liu2022devil}. 

This paper focuses on bridging this gap. We propose \name, a framework that can both generate realistic images and extract high-quality representations from images. Besides using variable masking ratio during pre-training, unlike previous MIM methods whose inputs are pixels, both the inputs and the reconstruction targets of \name~are \emph{semantic tokens}. This design improves both generation and representation learning, overcoming the issue described above. For generation, as shown in \autoref{fig:teaser}, operating in token space not only allows \name~to perform image generation tasks iteratively (\autoref{sec:evaluation}), but also enables \name~to learn a probability distribution of the masked tokens instead of an average of all possible masked pixels, leading to diverse generation results. For representation learning, using tokens as inputs and outputs allows the network to operate at a high semantic level \red{without losing low-level details}, leading to significantly higher linear probing performances than existing MIM methods.

We evaluate \name~on multiple generative and representation downstream tasks. As shown in \autoref{fig:lp-gen}, for class-\textbf{un}conditional image generation on ImageNet-1K, our method surpasses state of the art with both ViT-B and ViT-L (ViT-B achieves 11.11 FID \cite{heusel2017gans} and ViT-L achieves 9.10 FID), outperforming the previous state-of-the-art result by a large margin (MaskGIT \cite{chang2022maskgit} with 20.68 FID). This significantly push the limit of class-unconditional generation to a level even close to the state-of-the-art of class-conditional image generation ($\sim$6 FID \cite{chang2022maskgit, rombach2022high}), which is regarded as a much easier task in the literature \cite{luvcic2019high}. For linear probing on ImageNet-1K, our method with ViT-L achieves 78.9\% top-1 accuracy, surpassing all previous MIM-based representation learning methods and many strong contrastive baselines such as MoCo-v3 \cite{chen2021empirical}. Moreover, when combined with a simple contrastive loss \cite{simclr}, \name-C with ViT-L can further get 80.9\% accuracy, achieving state-of-the-art performance in self-supervised representation learning. We summarize our contributions:

\begin{Itemize}
    \item We introduce \name, a novel method that unifies generative model and representation learning by a single token-based MIM framework with variable masking ratios, introducing new insights to resolve the unification problem.
    \item \name~establishes a new state of the art on the task of class-unconditional image generation on ImageNet-1K.
    \item \name~further achieves state of the art in different downstream tasks, such as linear probing, few-shot learning, transfer learning, and class-conditional image generation.
\end{Itemize}
\section{Related Work}
\textbf{Self-supervised Learning in Computer Vision.}
Early work on unsupervised representation learning has focused on designing pretext tasks and training the network to predict their pseudo labels. Such tasks include solving jigsaw puzzles \cite{noroozi2016unsupervised}, restoring a missing patch \cite{pathak2016context}, or predicting image rotation \cite{gidaris2018unsupervised}. These pretext tasks result in representations that significantly trailed supervised training.

Contrastive learning \cite{chen2020simple,oord2018representation,chen2020big} has proven to be a competitive and systematic method to learn effective representations without human supervision, getting performance very close to that of supervised pre-training. A number of variants of contrastive learning have been proposed: SimCLR \cite{chen2020simple} uses a large batch size, and samples negative pairs within each batch; momentum-contrastive approach (MoCo) \cite{he2020momentum} leverages a moving-average encoder and a queue to store negative samples during training; Contrastive-Multiview-Coding \cite{tian2019contrastive} maintains a memory-bank to store features and generate negative samples. Some recent methods, like BYOL, do not rely on negative pairs \cite{chen2020exploring,grill2020bootstrap}. Instead, they use two neural networks that learn from each other to boost performance.

Recently, vision researchers have found that masked image modeling (MIM), modeled after techniques in NLP e.g. \cite{devlin2018bert}, is a very effective task for self-supervised learning. BEiT \cite{bao2021beit} recovers discrete visual tokens from masked inputs. PeCo \cite{dong2021peco} further regards MoCo-v3 \cite{chen2021empirical} as the perceptual model in VQGAN training to get a better tokenizer. MAE \cite{MAE} considers MIM as a denoising pixel-level reconstruction task, and CMAE \cite{CMAE} further combines MAE with a contrastive loss. Some other methods such as MaskFeat \cite{wei2022masked} and MVP \cite{wei2022mvp} predict features generated from teacher models.

However, current self-supervised learning methods based on MIM favor the performance of the representations on downstream tasks instead of the quality of the reconstructed images, leading to poor reconstructive results \cite{MAE, bao2021beit}. Our paper for the first time shows that a single model can not only learn high-level fine-grained representations, but also be used to generate images of high visual fidelity.

\textbf{Generative Models for Image Synthesis.} Recent years have witnessed tremendous progress in deep generative models for image synthesis. One major stream of generative models is built on top of generative adversarial networks (GANs) \cite{goodfellow2014generative, Han17, Karras2019, zhang2019self, brock2019large}. GAN-based models can generate realistic images in various domains, but suffer from training instabilities and mode collapse issues. Another stream is based on a two-stage scheme \cite{OordVK17, razavi2019generating, chang2022maskgit, yu2021vector, LeeKKCH22}: first tokenize the image into a latent space and then apply maximum likelihood estimation and sampling in the latent space. VQVAE-2 \cite{razavi2019generating} first shows this two-stage scheme can generate more diverse samples than GANs. ViT-VQGAN \cite{yu2021vector} uses ViT-based \cite{dosovitskiy2020image} encoder and decoder to get the latent code and then apply autoregressive generation in the latent space. MaskGIT \cite{chang2022maskgit} explores using a bidirectional transformer for token modeling and proposes parallel decoding for faster inference speed. Very recently, diffusion models~\cite{ho2020denoising, song2020score, dhariwal2021diffusion, rombach2022high} have also achieved superior results on image synthesis.

However, the above generative models lack the ability to extract high-quality semantic representations from images. Works such as \cite{donahue2019large} and \cite{yu2021vector} explore the possibility of using latent features as representations, but their performance is sub-optimal. Our method surpasses previous generative models on both class unconditional generation and representation learning by a large margin, showing that a unified, high-performance framework is feasible.


\begin{figure*}[t]
\begin{center}
\vspace{-10pt}
\includegraphics[width=1.0\textwidth]{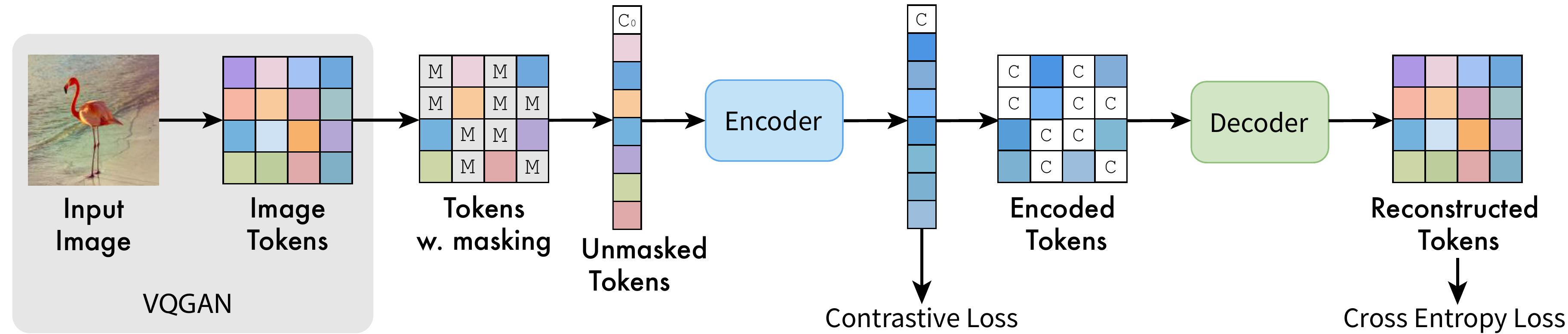}
\end{center}
\vspace{-15pt}
\caption{\name~Framework: we first use a VQGAN tokenizer to tokenize the input image into a sequence of semantic tokens. We then sample a masking ratio (see text for details on the sampling strategy) and randomly mask out tokens according to this sampled ratio. A ViT encoder-decoder structure processes the unmasked tokens. A reconstructive cross-entropy loss encourages the model to reconstruct masked tokens. We can also add an optional contrastive loss at the output of the encoder to further improve the linear separability of the learned latent feature space. }
\vspace{-10pt}
\label{fig:model}
\end{figure*}

\section{Method}

\name~is a unified framework for both generative tasks and representation learning. To enable such unification, we first use a pre-trained VQGAN model \cite{esser2021taming} to quantize input images into semantic tokens. Then we randomly mask out some input tokens using a variable masking ratio ranging from 0.5 to 1 (see \autoref{fig:model}), and apply an encoder-decoder transformer architecture on the remaining (unmasked) tokens to predict the masked tokens. We can further improve the separability of the learned representation by adding a simple yet effective contrastive loss similar to SimCLR \cite{simclr} on the output of the encoder (\name-C). Below, we describe our design in detail.

\subsection{Pre-training}

\textbf{Tokenization.} We first tokenize the input image into a sequence of semantic tokens using a tokenizer. The tokenizer employs the same setup as the first stage in the VQGAN model \cite{esser2021taming}. This tokenization step allows our model to operate on semantic tokens instead of raw pixels, which is beneficial for both generation and representation learning as shown in \autoref{fig:teaser} and \autoref{fig:lp-layer}.

\textbf{Masking Strategy.} To further bridge the gap between generative modeling and representation learning, we adopt a masking strategy with variable masking ratios. Specifically, we first randomly sample the masking ratio $m_r$ from a truncated Gaussian distribution centered at 0.55, left truncated by 0.5, and right truncated by 1. If the length of the input sequence of tokens is $l$, we randomly mask out $m_r\cdot l$ tokens and replace them with a learnable mask token \texttt{[M]} (\autoref{fig:model}). Since $m_r\geq 0.5$, we further randomly drop out $0.5\cdot l$ tokens from those masked tokens. Dropping a large fraction of masked tokens significantly reduces overall pre-training time and memory consumption, while helping both generation and representation performance. This is consistent with the findings in MAE \cite{MAE} for representation performance. 

\textbf{Encoder-Decoder Design.} After masking and dropping input tokens, following \cite{vit}, we concatenate a learnable ``fake" class token \texttt{[C$_0$]} to the input sequence. The concatenated sequence is then fed into a Vision Transformer (ViT) \cite{vit} encoder-decoder structure. Specifically, the ViT encoder takes the sequence of tokens after masking and dropping as input and encodes them into latent feature space. Before decoding, the output of the encoder is first padded to the full input length using the class token feature \texttt{[C]} learned by the encoder. As shown in MAE \cite{MAE}, the class token position can summarize global features of the input image. Thus, instead of using a learnable masking token
that is shared across different images, we use \texttt{[C]} that is specific to each image to pad the encoder outputs. We show in the Appendix that this design improves both generation and representation learning performance over using a masking token (as done in MAE \cite{MAE}). The decoder then takes the padded features to reconstruct the original tokens.

\textbf{Reconstructive Training.} Let $Y=[y_i]_{i=1}^N$ denote the latent tokens obtained from the tokenizer, where $N$ is the token sequence length, and $M=[m_i]_{i=1}^N$ denotes a corresponding binary mask determining which tokens are to be masked. The training objective is to reconstruct the masked tokens from the unmasked tokens. Therefore, we add a cross-entropy loss between the ground-truth one-hot tokens and the output of the decoder. Specifically,

\vspace{-5pt}
\begin{equation}
    \mathcal{L}_{reconstructive} = -\mathbb{E}_{Y\in \mathcal{D}} \big(\sum_{\forall i, m_i=1} \log p(y_i|Y_{M})\big),
\end{equation}
where $Y_M$ are the (subset of) \emph{unmasked} tokens in $Y$ and $p(y_i|Y_{M})$ is the probability predicted by the encoder-decoder network, conditioned on the unmasked tokens. Following MAE, we only optimize this loss on \emph{masked} tokens (optimizing the loss on all tokens reduces both generation and representation learning performance, similar to the observations in \cite{MAE}).

\textbf{Contrastive Co-training.} As shown in \cite{li2020making} and \cite{CMAE}, adding a contrastive loss in MIM method can further improve its representation learning performance. In our \name~framework, we can also add a contrastive loss to force better linear separability of the learned feature space. Similar to SimCLR \cite{chen2020big}, we add a two-layer MLP on top of the feature obtained by globally average pooling the output of the encoder. We then add an InfoNCE loss \cite{oord2018representation} on the output of the MLP head: 

\vspace{-5pt}
\begin{equation}
     \mathcal{L}_{contrastive} = -\frac{1}{B} \sum_{i=1}^B  \log \frac{e^{z_i^T\cdot z_i^+/\tau}}{ \sum\limits_{j=1}^B e^{z_i^T\cdot z_j/\tau}},
\end{equation}
where $z$ denotes the normalized features after the two-layer MLP, $B$ denotes the batch size, and $\tau$ denotes the temperature. The positive pairs $z_i, z_i^+$ are from two augmented views of the same image, and the negative samples $z_j$ are all other samples in the same batch. Our final loss is: 
\begin{equation}
    \mathcal{L} = \mathcal{L}_{reconstructive} + \lambda\cdot \mathcal{L}_{contrastive}
\end{equation} 
where $\lambda=0.1$ balances the scale of the two losses. We do not use the extensive augmentations typically used in contrastive learning, such as color jitter, random grey scale, or gaussian noise. This is because the reconstructive loss acts as a regularizer that prevents the encoder from learning shortcut solutions \cite{robinson2021can}. Our approach achieves superior performance on both generative tasks and representation learning even without the contrastive loss, and representation learning performance can be further boosted with the contrastive loss.

\subsection{Post-training Evaluation} 
\label{sec:evaluation}
To generate images for generative model evaluation, we use a \textbf{iterative decoding} strategy similar to MaskGIT \cite{chang2022maskgit}. We start from a blank image with all the tokens masked out. At each iteration, our model first predicts the tokens for the remaining masked tokens. Then we sample some of the predicted tokens (tokens that have a higher predicted probability are of higher probability to be sampled) and replace the corresponding masked tokens with these sampled predicted tokens. The number of masked tokens to be replaced in each iteration follows a cosine function, i.e., we replace fewer masked tokens in the early iterations and more masked tokens at later iterations. We use a total of 20 steps to generate an image. For representation learning, we globally average pool the features output from the ViT encoder, and use the pooled features as the input features for the classification head. A detailed description of our pre-training and evaluation implementations and architectures is provided in the Appendix.
\section{Results}
\name~is a unified framework for both generative model and representation learning. In this section, we conduct extensive experiments to evaluate the generation as well as visual representation capabilities.  To evaluate \name's generative performance, we conduct experiments on ImageNet-1K dataset \cite{russakovsky2015imagenet} for the task of class-unconditional image generation. To evaluate the quality of the learned representations,  we conduct experiments on ImageNet-1K dataset \cite{russakovsky2015imagenet} under two protocols: first is linear probing, where we add a linear classification head on top of the learned representations and only train the classification head, while keeping the backbone frozen; second is fine-tuning, where we fine-tune the whole parameters for the classification task. We also include results on few-shot learning and transfer learning to better evaluate the quality of the representations. More results and ablation studies can be found in the Appendix.

\subsection{Pre-training Setup}
We set the input image resolution as 256x256 to be consistent with previous generative models. After passing through the VQGAN tokenizer, the token sequence length is 16x16 (256 tokens). Following MAE \cite{MAE}, we use strong random crop and resize (0.2 to 1) and random flipping as our default augmentations. We also trained models with a weaker version of random crop and resize (range from 0.8 to 1), which we call ``w.a." in the results. We pre-train base- and large-size vision Transformers \cite{vit}, i.e., ViT-B and ViT-L, respectively. We use AdamW to train the model for 1600 epochs with batch size of 4096 for ViT-B, and batch size of 2048 for ViT-L. We use a cosine learning rate schedule with an 80-epoch warmup. The base learning rate is $1.5\times10^{-4}$ for both ViT-B and ViT-L, and is further scaled by batchsize/256. More details are in the Appendix.

\subsection{Image Generation}

\begin{table}[h]
\vspace{-10pt}
\caption{Quantitative comparison with state-of-the-art generative models on ImageNet 256x256 for class-unconditional generation. The number of parameters includes encoder, decoder, and detokenizer.}
\vspace{-15pt}
\label{tab:cls-uncond}
\begin{center}{
\resizebox{0.48\textwidth}{!}{
\begin{tabular}{l|c|cccccc}
\toprule
Methods & RES & FID$\downarrow$ & IS$\uparrow$ & \#params \\ 
\midrule
Self-Conditioned GAN \cite{liu2020diverse} & 128 & 40.3 & 15.82 & - \\
BigGAN \cite{donahue2019large} & 256 & 38.6 & 24.70 & $\sim$70M \\
BigGAN \cite{donahue2019large} & 128 & 30.9 & 23.56 & $\sim$70M \\
BigGAN+Clustering \cite{luvcic2019high} & 128 & 22.0 & 23.5 & $\sim$70M \\
HiT \cite{zhao2021improved} & 128 & 30.8 & 21.64 & $\sim$30M \\
ADM \cite{dhariwal2021diffusion} & 256 & 26.2 & 39.70 & 554M \\
MaskGIT \cite{chang2022maskgit} & 256 & 20.7 & 42.08 & 203M \\
\midrule
\name~(ViT-B) & 256 & 11.1 & 81.17 & 176M \\
\name~(ViT-B, w.a.) & 256 & \textbf{8.67} & \textbf{94.8} & 176M \\
\name~(ViT-L) & 256 & 9.10 & 105.1 & 439M \\
\name~(ViT-L, w.a.) & 256 & \textbf{7.04} & \textbf{123.5} & 439M \\
\bottomrule
\end{tabular}
}}
\end{center}
\vspace{-10pt}
\end{table}

\textbf{Class-Unconditional Image Generation.}
Our pre-trained model can naturally perform class-unconditional image generation without any fine-tuning on the model parameters. \autoref{tab:cls-uncond} compares the class-unconditional image generation results of our model and SOTA generative models on ImageNet, reporting Frechet Inception Distance (FID) \cite{heusel2017gans} and Inception Score (IS) \cite{salimans2016improved} as standard metrics. As shown in the table, our method outperforms all previous image generation methods by a large margin. The previous SOTA can only achieve $20.7$ FID and $42.08$ IS, while our ViT-B model can achieve $11.1$ FID and $81.17$ IS with similar number of parameters. \red{This is likely because our framework can extract much better representations than all previous generative models as shown in \autoref{tab:linear-probe}, leading to superior generative performance}. Our ViT-L model further achieves $9.10$ FID and $7.04$ FID when trained with weak augmentation, which is very close to the \emph{class-conditional} generation performance of transformer models (e.g. $6.18$ FID in MaskGIT \cite{chang2022maskgit}), a much easier task than class-unconditional generation \cite{luvcic2019high}.

\begin{figure}[t]
\vspace{-15pt}
\begin{center}
\includegraphics[width=0.235\textwidth]{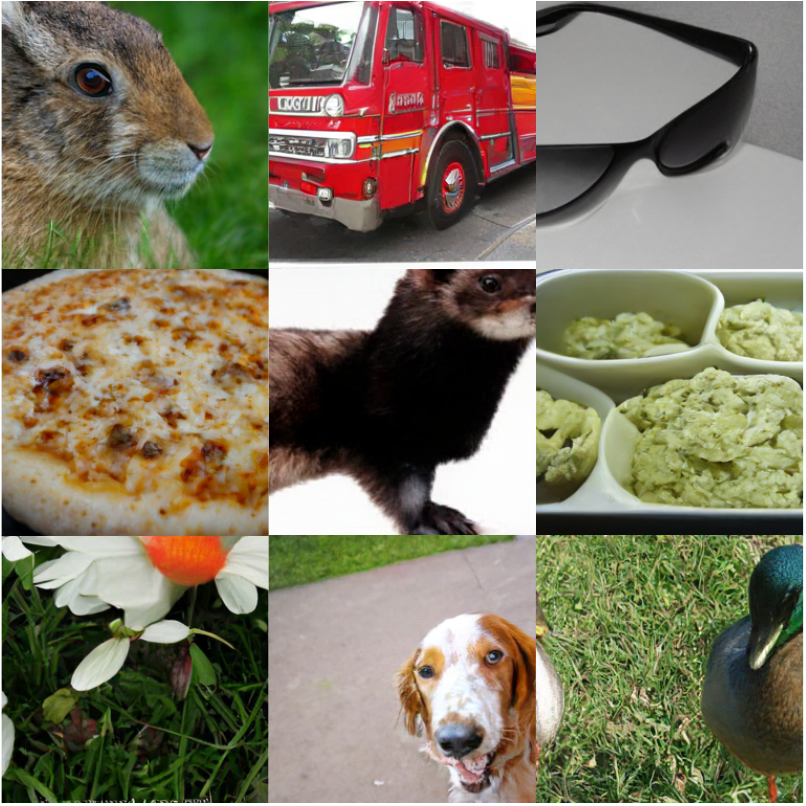}
\includegraphics[width=0.235\textwidth]{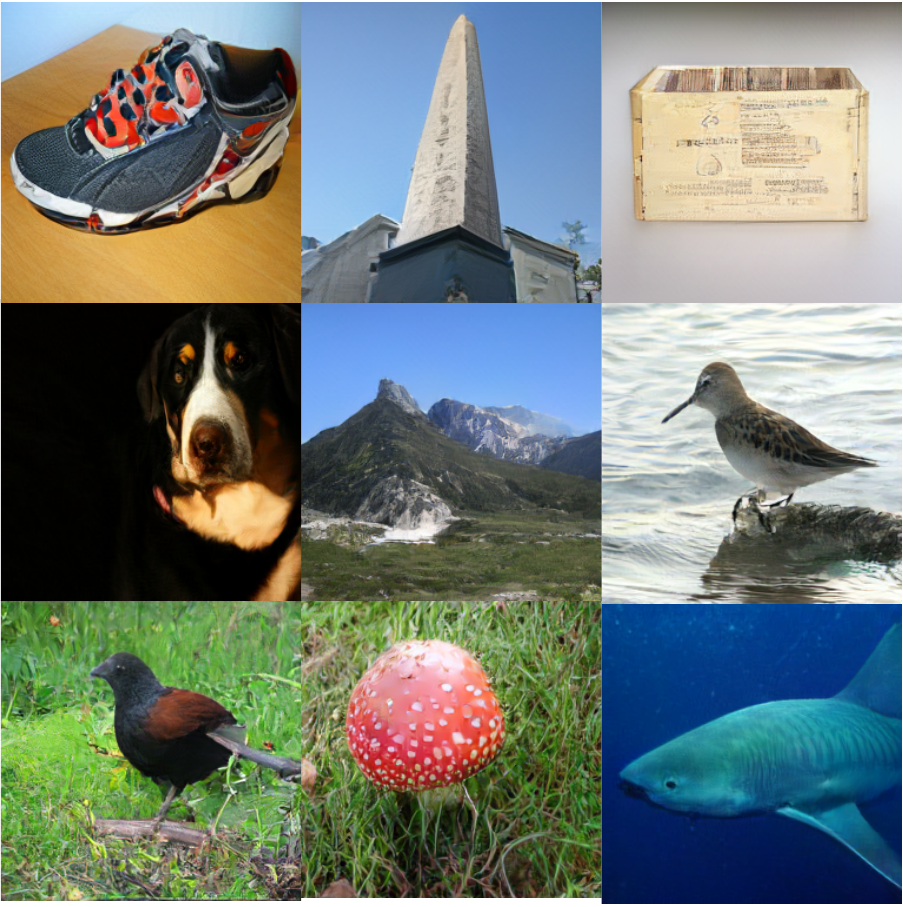}
{(a) Default augmentation}  \hspace*{0.25in}  {(b) Weak augmentation}
\end{center}
\vspace{-10pt}
\caption{Images generated by \name~(ViT-L). (a) images generated from \name~trained with default strong augmentation, i.e., crops out larger portion of the image. (b) images generated from \name~trained with weak augmentations, i.e., crops out smaller portion of the image. We see that visual fidelity and diversity are very good for both models.} 
\vspace{-10pt}
\label{fig:cls-uncond}
\end{figure}


We also note that the augmentations used to train the tokenizer and the encoder-decoder model can affect the evaluation scores. As shown in \autoref{tab:cls-uncond} and \autoref{fig:cls-uncond}, with default ``strong" augmentation (i.e. random resized crop scale from 0.2 to 1), the FID and IS of the model are worse than using a weaker augmentation (random resized crop scale from 0.8 to 1). One possible reason is that the ImageNet validation set used to compute the FID is resized to 256 and center cropped. Since FID is computed based on the similarity between the generated image and the images in ImageNet validation set, it will be higher if the scale of the generated image is smaller. However, this does not necessarily mean that the visual quality of the generated image is worse. As shown in \autoref{fig:cls-uncond}, images generated with default augmentation can be much more zoomed in and off-center, but the images are still realistic and of high quality. We include more results in the Appendix, including for class conditional generation and image editing tasks such as inpainting.

\begin{table}[t]
\caption{Top-1 accuracy of linear probing on ImageNet-1k. $^\dag$ denotes methods which require additional teacher model (CLIP) trained from image-text data. $^*$ denotes methods using multi-crop augmentations. RN is short for ResNet. The number of parameters for MAGE includes VQ-GAN tokenizer and ViT encoder.
}
\vspace{-13pt}
\label{tab:linear-probe}
\begin{center}{
\resizebox{0.48\textwidth}{!}{
\begin{tabular}{l|ccccccc}
\toprule
Methods & Model & \#params  & Acc. \\ 
\midrule
\textit{ Generative models} \\
BigBiGAN \cite{donahue2019large} & RN50 & 23M  & 56.6 \\
MaskGIT \cite{chang2022maskgit} & BERT & 227M  & 57.4 \\
ViT-VQGAN \cite{yu2021vector} & VIM-Base & 650M  & 65.1 \\
ViT-VQGAN \cite{yu2021vector} & VIM-Large & 1697M  & 73.2 \\
\midrule
\textit{ MIM methods} \\
BEiT \cite{bao2021beit} & ViT-B & 86M  & 56.7 \\
MAE \cite{MAE} & ViT-B & 86M  & 68.0 \\
Ge$^2$-AE \cite{liu2022devil} & ViT-B & 86M  & \textbf{75.3} \\
\rowcolor{LightCyan}
\name & ViT-B & 24M+86M  & 74.7 \\
\midrule
MAE \cite{MAE} & ViT-L & 304M  & 75.8 \\
\rowcolor{LightCyan}
\name & ViT-L & 24M+304M & \textbf{78.9} \\
\midrule
\textit{ Contrastive methods} \\
SimCLRv2 \cite{chen2020big} & RN50w2 & 94M  & 75.6 \\
BYOL \cite{grill2020bootstrap} & RN50w2 & 94M  & 77.4 \\
CAE \cite{chen2022context} & ViT-B & 86M  & 70.4 \\
CMAE \cite{CMAE} & ViT-B & 86M  & 73.9 \\
MoCo v3 \cite{chen2021empirical} & ViT-B & 86M  & 76.7 \\
DINO \cite{zhou2021ibot} & ViT-B & 86M  & 72.8 \\
iBOT \cite{zhou2021ibot} & ViT-B & 86M  & 76.0 \\
\rowcolor{LightCyan}
\name-C & ViT-B & 24M+86M  & \textbf{78.2} \\
\midrule
SimCLRv2 \cite{chen2020big} & RN152w2 & 233M  & 77.4 \\
BYOL \cite{grill2020bootstrap} & RN200w2 & 250M  & 79.6 \\
MoCo v3 \cite{chen2021empirical} & ViT-L & 304M  & 77.6 \\
CAE \cite{chen2022context} & ViT-L & 304M  & 78.1 \\
MoCo v3 \cite{chen2021empirical} & ViT-H & 632M  & 78.1 \\
\rowcolor{LightCyan}
\name-C & ViT-L & 24M+304M  & \textbf{80.9} \\
\midrule
\textit{ Additional Data/Aug.} \\
MVP$^\dag$ \cite{wei2022mvp} & ViT-B & 86M  & 75.4 \\
BEiT v2$^\dag$ \cite{peng2022beit} & ViT-B & 86M  & 80.1 \\
SwAV$^*$ \cite{caron2020unsupervised} & RN50w5 & 586M  & 78.5 \\
DINO$^*$ \cite{caron2021emerging} & ViT-B & 86M  & 78.2 \\
iBOT$^*$ \cite{zhou2021ibot} & ViT-B & 86M  & 79.5 \\
iBOT$^*$ \cite{zhou2021ibot} & ViT-L & 304M & 81.0 \\
\bottomrule
\end{tabular}
}}
\end{center}
\vspace{-20pt}
\end{table}

\subsection{Image Classification}

\textbf{Linear Probing.}
Linear probing is a primary evaluation protocol for self-supervised learning. As shown in \autoref{tab:linear-probe}, \name~outperforms MAE \cite{MAE} by 6.7\% on ViT-B and 3.1\% on ViT-L for ImageNet-1K linear probe top-1 accuracy, achieving state-of-the-art results among all MIM methods. Moreover, a simple contrastive loss similar to SimCLR \cite{chen2020simple} can further boost our performance. We do not use color jitter, random grey scale, or multi-crop augmentations used in SwAV \cite{caron2020unsupervised}, DINO \cite{caron2020unsupervised} and iBOT \cite{zhou2021ibot}. Multi-crop augmentation typically brings 3\%-5\% improvements on accuracy, but introduces large computational overheads. In spite of no multi-crop, \name-C achieves $78.2\%$ accuracy with ViT-B and $80.9\%$ accuracy with ViT-L. Our ViT-B performance surpasses that of ViT-H in MoCo v3 (632M parameters), indicating that the extra parameters (24M) in the tokenizer are \emph{not} the reason for our good performance.

\begin{table}[t]
\caption{Few-shot evaluation on ImageNet-1K. We report the top-1 accuracy with different self-supervised methods and different numbers of the ImageNet-1K labels used. We report the accuracy of MAE under our implementation (denoted by $\dag$). Note that MSN \cite{assran2022masked} uses multi-crop augmentation.
}
\vspace{-15pt}
\label{tab:few-shot}
\begin{center}{
\begin{tabularx}{0.46\textwidth}{l|>{\centering}X
>{\centering}X
>{\centering}X
>{\centering\arraybackslash}X}
\toprule
\multirow{2}{*}{Method} & \multicolumn{4}{c}{Training images per ImageNet Class} \\ 
     &  5 & 10 & 13 & 25 \\

\midrule
\textit{ ViT-B} \\
MAE$^\dag$ \cite{MAE} & 29.2 & 34.5 & - & 38.7\\
MSN \cite{assran2022masked} & \textbf{65.5} & - & \textbf{69.6} & -\\

\rowcolor{LightCyan}
\name & 53.5 & 58.4 & 59.7 & 61.7 \\
\rowcolor{LightCyan}
\name-C  & 62.7 & 66.9 & 67.8 & 69.1 \\

\midrule
\textit{ ViT-L} \\
MAE$^\dag$ \cite{MAE} & 42.2 & 47.7 & - & 51.7 \\
MSN \cite{assran2022masked} & - & - & 70.1 & -\\
\rowcolor{LightCyan}
\name & 60.3 & 66.1 & 67.8 & 69.6 \\
\rowcolor{LightCyan}
\name-C & \textbf{68.1} & \textbf{71.9} & \textbf{73.0} & \textbf{74.2} \\

\bottomrule

\end{tabularx}
}
\end{center}
\vspace{-15pt}
\end{table}

\begin{figure*}[h]
\begin{center}
\vspace{-10pt}
\includegraphics[width=1\linewidth]{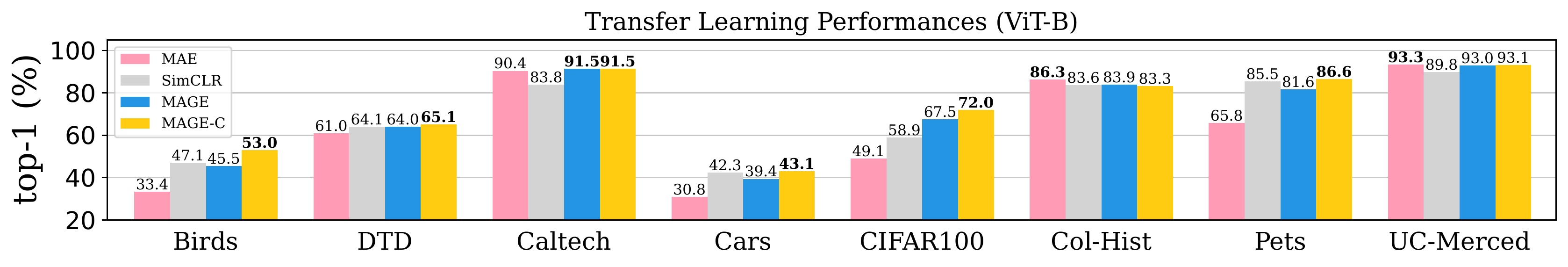}

\includegraphics[width=1\linewidth]{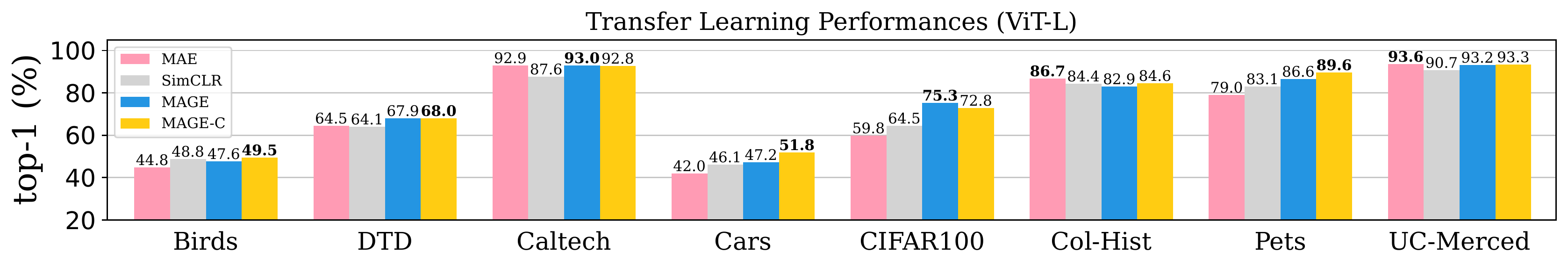} 

\vspace{-18pt}
\end{center}
\caption{Transfer learning performance of ViT-B and ViT-L pre-trained on ImageNet-1K using different methods. Our methods outperforms SimCLR \cite{simclr} and MAE \cite{MAE} on 6 of the 8 datasets.}\label{fig:transfer}
\vspace{-15pt}
\end{figure*}

\textbf{Few-shot Learning.}
The premise of self-supervised learning is to learn representations on unlabeled data that can be effectively applied to prediction tasks with few labels \cite{chen2020big}. Following \cite{dosovitskiy2020image}, we freeze the weights of the pre-trained model and train a linear classifier on top using a few labeled samples. As shown in \autoref{tab:few-shot}, our methods with ViT-B outperform MAE \cite{MAE} by a very large margin and achieves similar performance as MSN \cite{assran2022masked}, which is the state-of-the-art method for self-supervised label-efficient learning. Moreover, the performance of \name-C with ViT-L even surpasses the performance of MSN using 13 images per class (1\% of ImageNet-1K), even though MSN uses multi-crop augmentation.

\textbf{Transfer Learning.}
Another important property of self-supervised representation is its transferability to different datasets. Following the protocol in \cite{dosovitskiy2020image}, we evaluate the transfer learning performance of \name~pre-trained on ImageNet-1K on $8$ datasets under a few-shot setting (25 samples per class). Results are shown in \autoref{fig:transfer}: we see that \name's superior performance on ImageNet-1K translates to strong performance on other tasks. Since our method operates on quantized semantic tokens instead of raw pixels, it is likely to be more robust to domain shift.

\begin{table}[t]
\caption{Fine-tuning performance on ImageNet-1K. We report the top-1 accuracy and the improvement over training-from-scratch for different methods (other numbers taken from the respective papers). The ViT models trained from scratch on semantic tokens follow the exact same training setting as the ViT models trained from scratch on original image pixels in \cite{MAE}.}
\vspace{-15pt}
\label{tab:finetune}
\begin{center}{
\begin{tabular}{l|llllll}
\toprule
Method  & ViT-B & ViT-L  \\

\midrule
\rowcolor{Grey}
scratch on pixels & 82.3 & 82.6 \\
DINO \cite{caron2021emerging} & 82.8 \textcolor{darkgreen}{(+0.5)} & - \\
MoCo v3 \cite{chen2021empirical} & 83.2 \textcolor{darkgreen}{(+0.9)} & 84.1 \textcolor{darkgreen}{(+1.5)} \\
BEiT \cite{bao2021beit} & 83.2 \textcolor{darkgreen}{(+0.9)} & 85.2 \textcolor{darkgreen}{(+2.6)} \\
MAE \cite{MAE} & 83.6 \textcolor{darkgreen}{(+1.3)} & 85.9 \textcolor{darkgreen}{(+3.3)} \\
CAE \cite{chen2022context} & 83.9 \textcolor{darkgreen}{(+1.6)} & 86.3 \textcolor{darkgreen}{(+3.7)} \\ 
MVP \cite{wei2022mvp} & 84.4 \textcolor{darkgreen}{(+2.1)} & 86.3 \textcolor{darkgreen}{(+3.7)} \\
PeCo \cite{dong2021peco} & 84.5 \textcolor{darkgreen}{(+2.2)} & 86.5 \textcolor{darkgreen}{(+3.9)} \\
\midrule
\rowcolor{Grey}
scratch on tokens & 80.7 & 80.9 \\
\rowcolor{LightCyan}
\name & 82.5 \textcolor{darkgreen}{(+1.8)} & 83.9 \textcolor{darkgreen}{(+3.0)}\\
\rowcolor{LightCyan}
\name-C & 82.9 \textcolor{darkgreen}{(+2.2)} & 84.3 \textcolor{darkgreen}{(+3.4)} \\

\bottomrule

\end{tabular}
}
\end{center}
\vspace{-15pt}
\end{table}

\textbf{Fine-tuning.}
\autoref{tab:finetune} shows the fine-tuning performance of \name~and other self-supervised learning methods, when we can change all the pre-trained encoder parameters. Our method achieves performance at par with DINO \cite{caron2020unsupervised} and slightly under MoCo-v3 \cite{chen2021empirical}. We believe that the use of quantized tokens leads to a subpar from-scratch and fine-tune performance, and leave further investigations of this phenomenon to future work. We note, however, that our method still improves over our supervised training-from-scratch baseline by as large a margin as other methods.

\begin{table*}[t]
\caption{Top-1 accuracy of linear probing and class unconditional generation FID of \name~on ImageNet-1k with different masking ratio distribution. $\mu$ denotes the mode and $\sigma$ the standard deviation of the truncated Gaussian distribution. When $\sigma=0$, the masking ratio is fixed and generation has poor quality with very high FID ($>$50). Therefore we put N/A for FID in such cases.}
\vspace{-15pt}
\label{tab:mask-ratio}
\begin{center}{
\begin{tabular}{l|ccccc|cccc}
\toprule
& \begin{tabular}[c]{@{}c@{}} $\mu=0.7$ \\ $\sigma=0$ \end{tabular}
& \begin{tabular}[c]{@{}c@{}} $\mu=0.6$ \\ $\sigma=0$ \end{tabular}
& \begin{tabular}[c]{@{}c@{}} $\mu=0.55$ \\ $\sigma=0$ \end{tabular}
& \begin{tabular}[c]{@{}c@{}} $\mu=0.5$ \\ $\sigma=0$ \end{tabular}
& \begin{tabular}[c]{@{}c@{}} $\mu=0.45$ \\ $\sigma=0$ \end{tabular}
& \begin{tabular}[c]{@{}c@{}} $\mu=0.55$ \\ $\sigma=0$ \end{tabular}
& \begin{tabular}[c]{@{}c@{}} $\mu=0.55$ \\ $\sigma=0.15$ \end{tabular}
& \begin{tabular}[c]{@{}c@{}} $\mu=0.55$ \\ $\sigma=0.25$ \end{tabular}
& \begin{tabular}[c]{@{}c@{}} $\mu=0.55$ \\ $\sigma=0.5$ \end{tabular} \\ 

\midrule
Linear Probing & 69.7 & 70.1  & 71.5 & 70.9 & 70.4 &  71.5 & 72.0 & \textbf{72.2} & 71.8 \\
FID & N/A & N/A & N/A & N/A & N/A & N/A  & 12.5 & \textbf{12.2} & 13.0 \\
\bottomrule
\end{tabular}
}
\end{center}
\vspace{-18pt}
\end{table*}

\subsection{Analysis}

In this section, we analyze the two key components of \name~that enables the unification of generative modeling and representation learning: variable masking ratio and quantized tokenization. All experiments are conducted on ViT-B. Experiments on variable masking ratio are all trained for 400 epochs, and experiments on quantized tokenization are all trained for 1600 epochs. More analysis and ablation studies are in the Appendix.

\textbf{Masking Design.} Variable masking ratio is one of our key components. We find that the quality of the learned representation can also be affected by the distribution used to sample our masking ratio. We compare the results of \name~on linear probing and class unconditional generation on ImageNet-1k using different masking ratio distributions in \autoref{tab:mask-ratio}. We denote the mode of the truncated Gaussian distribution as $\mu$, and the standard deviation of the truncated Gaussian distribution as $\sigma$. Note that $\sigma=0$ indicates a fixed masking ratio. The left 5 columns ablate $\mu$, and the right 4 columns ablate $\sigma$. The results show that a variable masking ratio is necessary to enable generation. Moreover, using a variable masking ratio also enables representation learning to learn better features and achieve better linear probe performance.

\begin{table}[t]
\caption{Reconstruction loss and linear probe accuracy of \name~with unquantized features and quantized tokens as input. Using unquantized features makes it much easier to infer masked tokens, and hence results in worse linear probe performance.}
\vspace{-10pt}
\label{tab:analysis}
\begin{center}{
\begin{tabular}{c|ccc}
\toprule
inputs  & recon. loss & linear probe (\%)  \\

\midrule

Unquantized features & 3.31 & 49.5 \\
Quantized tokens  & 5.76 & 74.7 \\

\bottomrule
\end{tabular}
}
\end{center}
\vspace{-15pt}
\end{table}

\begin{figure}[t]
\begin{center}
\includegraphics[width=1.0\linewidth]{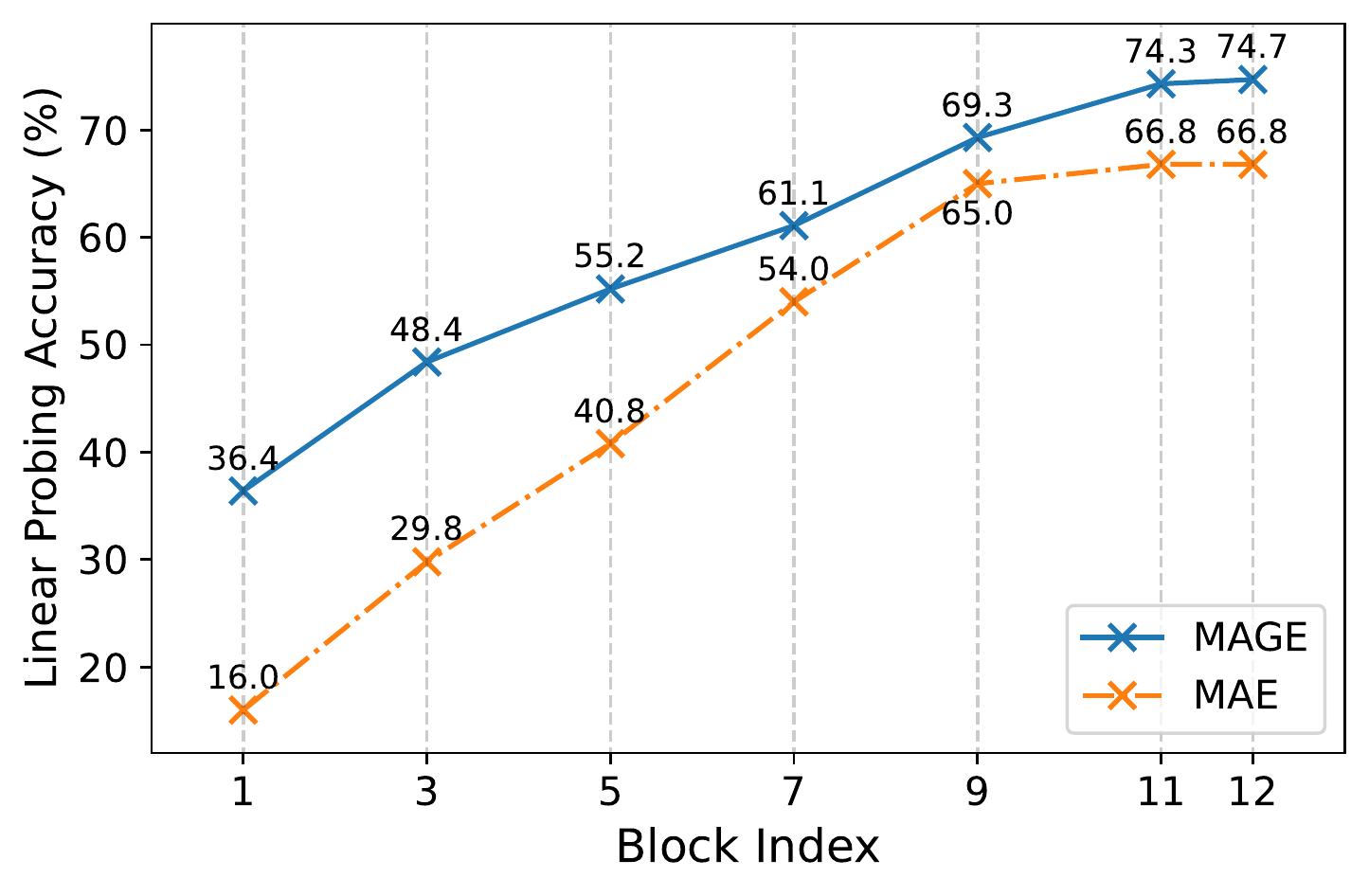}

\vspace{-15pt}
\end{center}
\caption{Linear probe accuracy of MAE and \name~at different transformer blocks of ViT-B. \name~ consistently has higher accuracy across all transformer blocks due to the semantic nature of the quantized tokens.}\label{fig:lp-layer}
\vspace{-15pt}
\end{figure}

\textbf{Tokenization.}
Previous self-supervised learning methods on images typically directly use raw images as the inputs of the transformer. However, in \name~we use quantized semantic tokens as both inputs and reconstruction targets. We elaborate the benefits of this design as follows:
\begin{Itemize}
    \item First, during generation, it allows the network to iteratively use its output as the input in the next iteration, which enables high-quality and diverse image reconstruction and generation, as shown in \autoref{fig:teaser} and \autoref{fig:cls-uncond}. 
    \item Second, it allows the whole network to operate at a semantic level without losing low-level details and thus extract better representations. We demonstrate this by comparing the linear probe performance on features from each transformer block of ViT-B trained using MAE and \name. As shown in \autoref{fig:lp-layer}, the linear probe accuracy of \name~at each transformer block is always higher than MAE throughout the encoder.
    \item  Third, the quantizer prevents shortcuts created by the VQGAN CNN encoder. If we directly use features extracted by the VQGAN encoder without quantization as input to the transformer, since the receptive fields of neighboring feature pixels have significant overlap, it is much easier to infer masked feature pixels using nearby unquantized feature pixels. As shown in \autoref{tab:analysis}, with the same masking strategy, using unquantized features achieves much lower reconstructive loss ($3.31$ vs. $5.76$), but also a much lower linear probe accuracy ($49.5\%$ vs. $74.7\%$). This suggests that the pre-training task is too easy, leading to shortcut solutions, and hence to poor representations. The quantization step is therefore necessary to learn good representations.
\end{Itemize}

\section{Discussion}
We have presented MAGE, a masking-based approach that unifies image generation and representation learning in a simple and effective framework. The key to our method is the use of quantized tokens and the use of variable masking ratios that adapt smoothly to both tasks (generation and representation). We have shown extensive results on linear probing, few-shot transfer learning, and unconditional image generation. To the best of our knowledge, this is the first model that achieves close to SOTA results for both tasks using the same data and training paradigm. A natural future extension is to pre-train on larger unlabeled datasets such as JFT300 to further improve performance.
\section{Acknowledgement}
We thank Google Cloud Platform (GCP) and MIT-IBM Watson AI Lab for providing necessary computation resources for this project.


{\small
\bibliographystyle{configs/ieee_fullname}
\bibliography{main.bib}
}

\clearpage

\appendix

\twocolumn[{
\renewcommand\twocolumn[1][]{#1}

\begin{center}
  \centering
\includegraphics[width=1\linewidth]{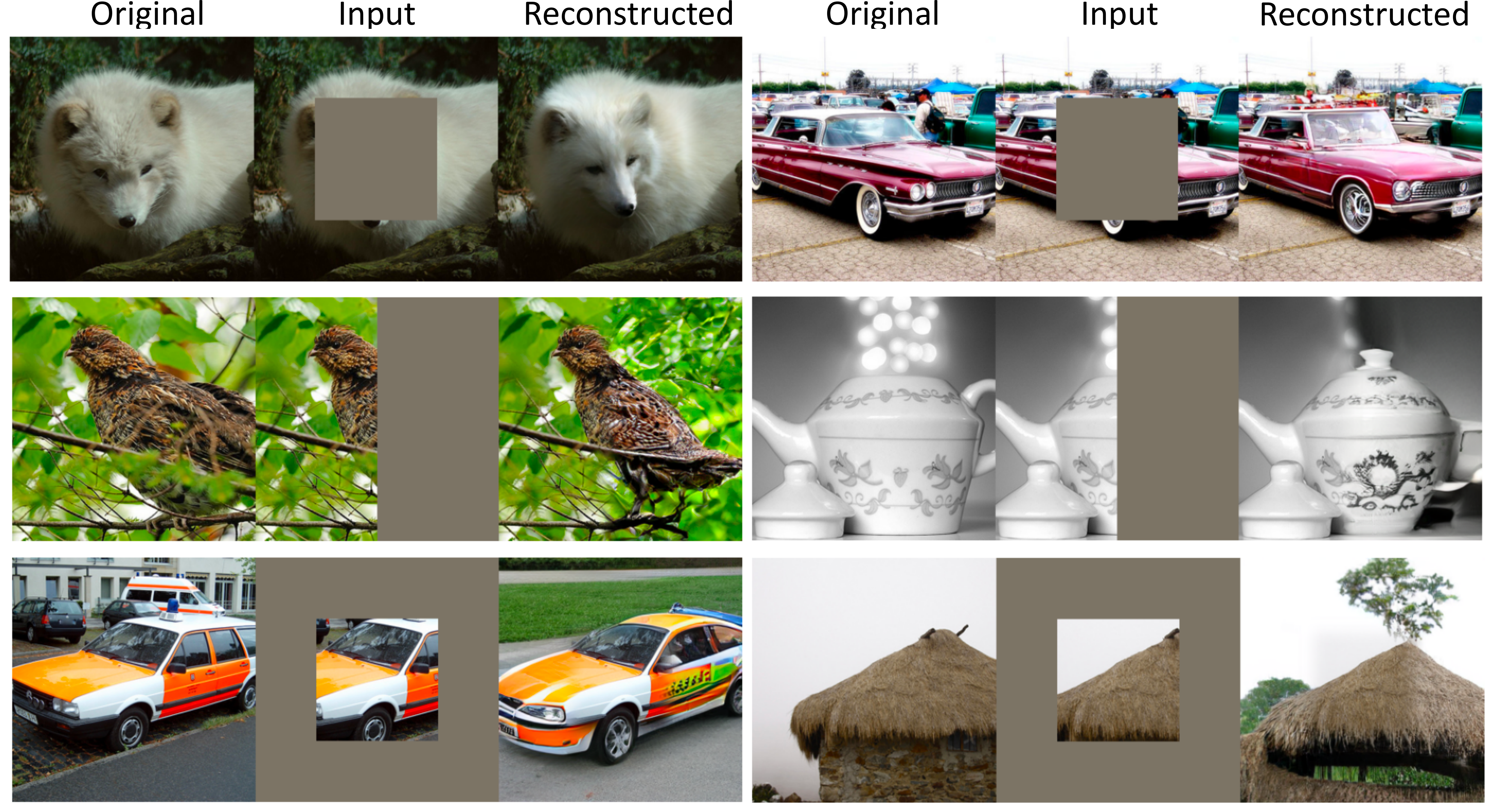}
  \captionof{figure}{Examples of image inpainting (first row), outpainting (second row), and uncropping (outpainting on a large mask, third row) using \name~(ViT-L).}\label{fig:qualitative}
\end{center}
}]

\section{Additional Results}
\subsection{Qualitative Results}

\begin{figure*}[t]
\begin{center}

\includegraphics[width=0.9\linewidth]{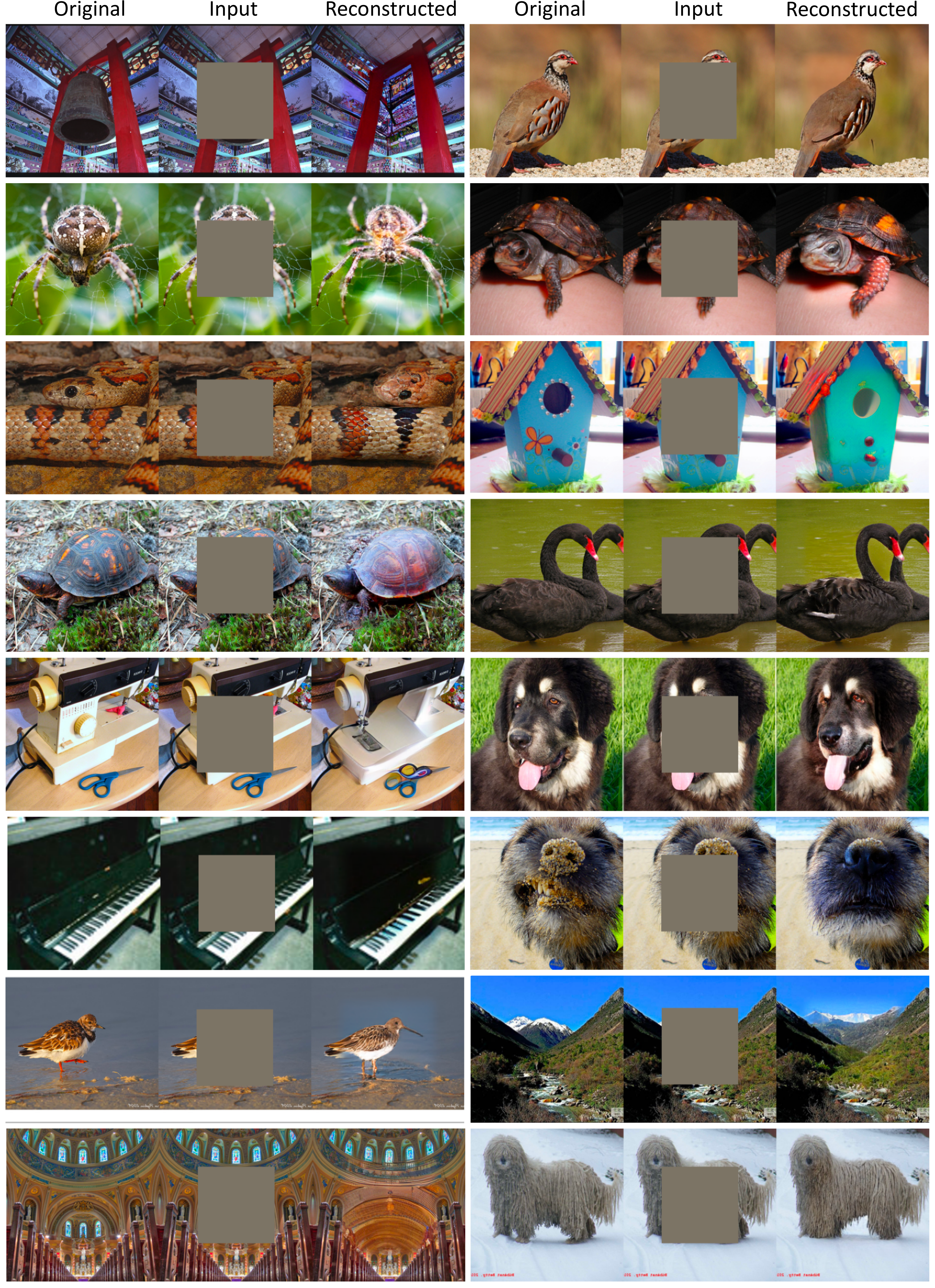}

\end{center}
\caption{More examples of image inpainting using \name~ (ViT-L).}\label{fig:inpainting-gallery}
\vspace{-15pt}
\end{figure*}

\begin{figure*}[t]
\begin{center}

\includegraphics[width=0.9\linewidth]{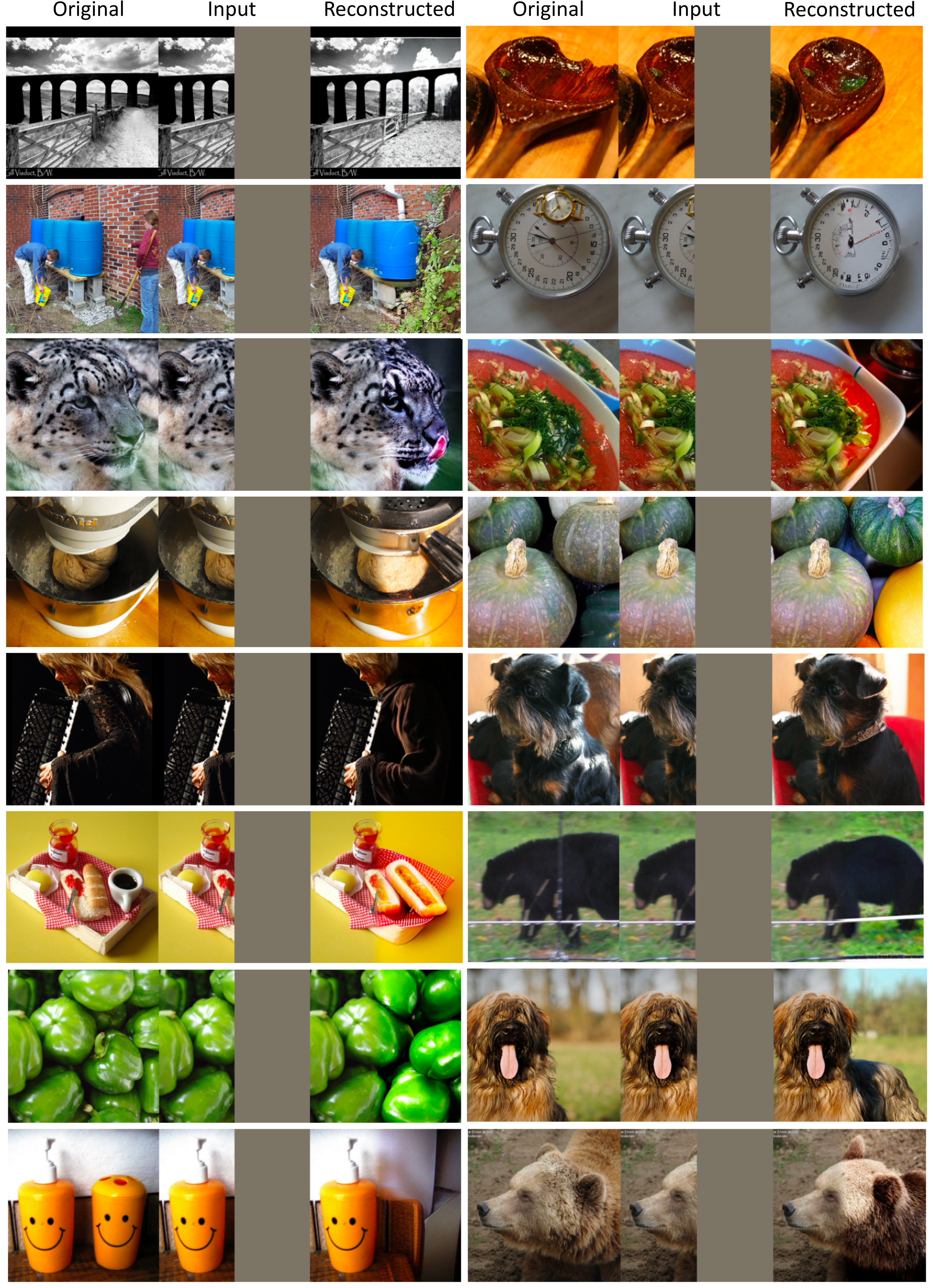}

\end{center}
\caption{More examples of image outpainting using \name~ (ViT-L).}\label{fig:outpainting-gallery}
\vspace{-15pt}
\end{figure*}

\begin{figure*}[t]
\begin{center}

\includegraphics[width=0.9\linewidth]{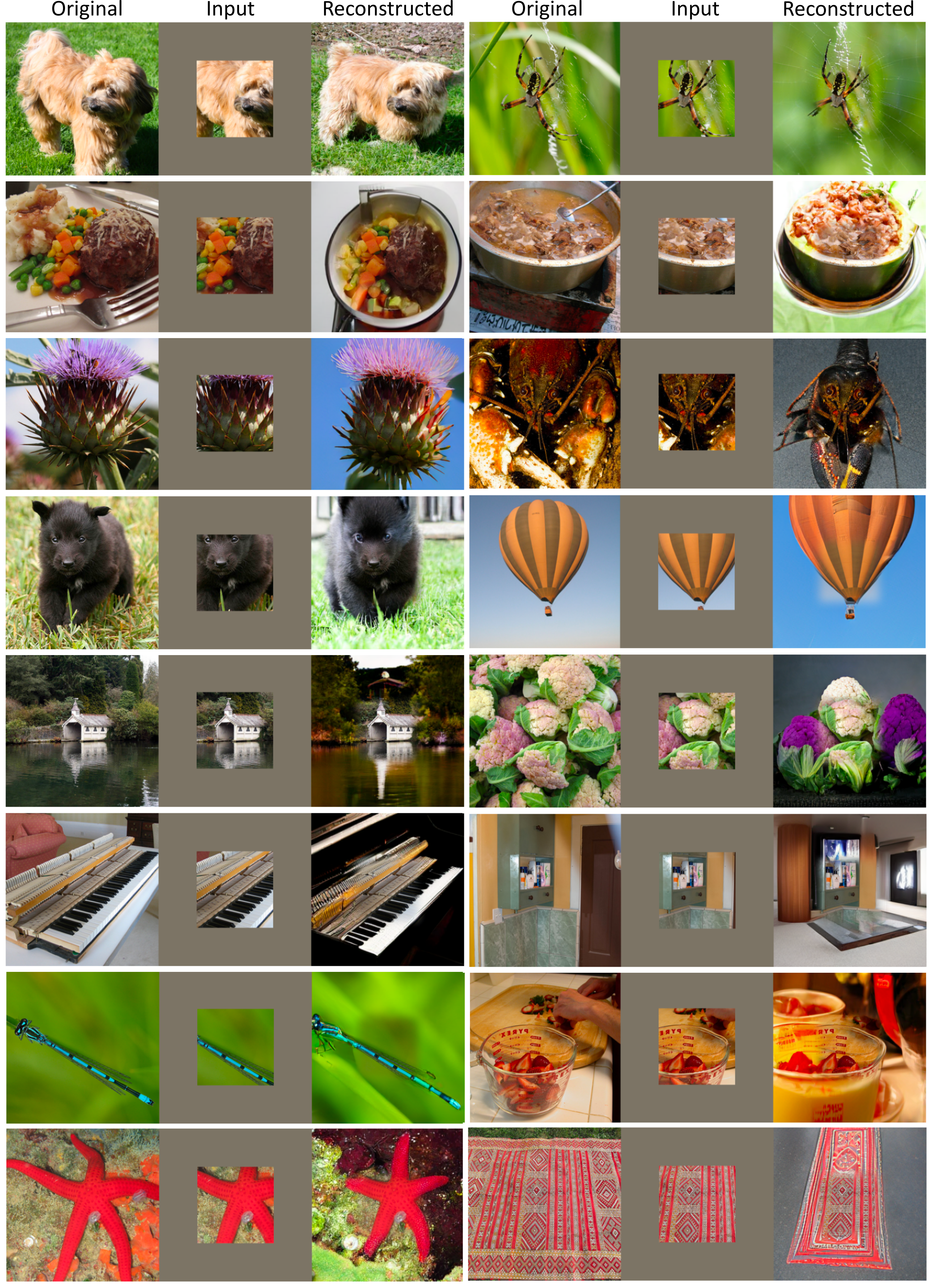}

\end{center}
\caption{More examples of image outpainting on large outpainting mask (uncropping) using \name~(ViT-L).}\label{fig:uncropping-gallery}
\vspace{-15pt}
\end{figure*}

\textbf{Image Inpainting and Outpainting.} With the superior class-unconditional reconstruction and generation power shown in the main paper, \name~naturally enables many common image synthesis applications. As shown in \autoref{fig:qualitative}, \name~can reconstruct realistic and high-quality images for different image editing tasks such as inpainting (first row), outpainting (second row), and uncropping (outpainting on large masks, third row). We also include more qualitative results in \autoref{fig:inpainting-gallery}, \autoref{fig:outpainting-gallery}, and \autoref{fig:uncropping-gallery}, demonstrating \name's excellent ability in such image synthesis tasks. All results are generated using \name~
based on ViT-L trained with default augmentations, and the original images are all from the ImageNet eval set.

\begin{figure*}[t]
\begin{center}

\includegraphics[width=0.8\linewidth]{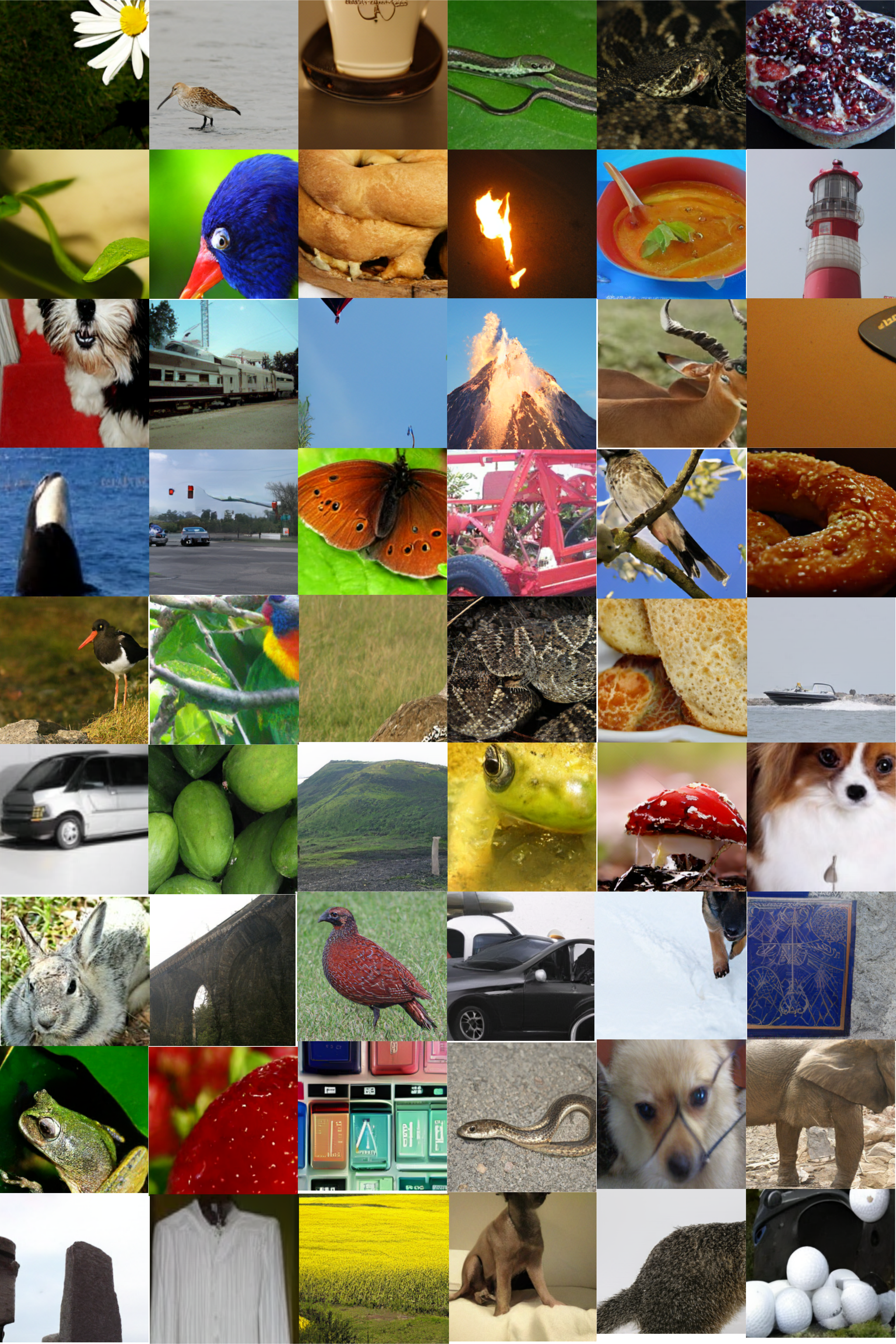}

\end{center}
\caption{More uncurated examples of Class-unconditional image generation on ImageNet using \name~trained with default strong augmentation.}\label{fig:cls-uncond-strong-gallery}
\vspace{-15pt}
\end{figure*}

\begin{figure*}[t]
\begin{center}

\includegraphics[width=0.8\linewidth]{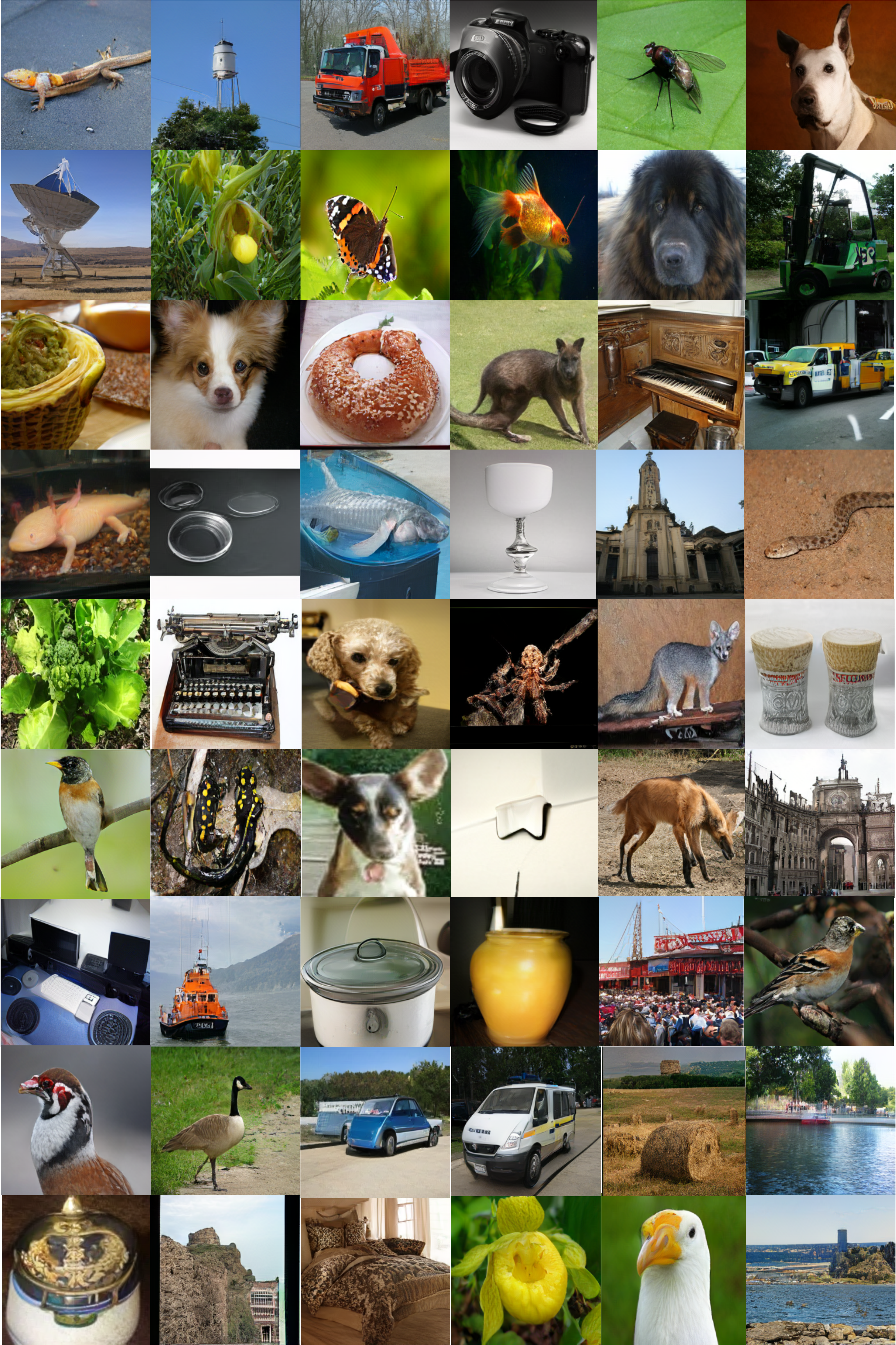}

\end{center}
\caption{More uncurated examples of Class-unconditional image generation on ImageNet using \name~trained with weak augmentation.}\label{fig:cls-uncond-weak-gallery}
\vspace{-15pt}
\end{figure*}

\textbf{Class Unconditional Generation.} We include more class unconditional generation results using default strong augmentation (random crop and resize (0.2 to 1) and random flipping) and weak augmentation (random crop and resize (0.8 to 1) and random flipping) in \autoref{fig:cls-uncond-strong-gallery} and \autoref{fig:cls-uncond-weak-gallery}.

\subsection{Quantitative Results}

\begin{table}[h]
\caption{Quantitative comparison with state-of-the-art generative models on ImageNet 256x256 for class-conditional generation. Our method uses a \name~pre-trained ViT-B as encoder and only trains a class-conditional decoder with 113M parameters.}
\vspace{-10pt}
\label{tab:cls-cond}
\begin{center}{
\begin{tabular}{l|ccccccc}
\toprule
Methods  & FID$\downarrow$ & IS$\uparrow$ & \#params \\ 
\midrule

DCTransformer \cite{nash2021generating} & 36.51 & - & 738M \\
VQVAE-2 \cite{razavi2019generating} & 31.11 & $\sim$45 & 13.5B \\
VQGAN \cite{chang2022maskgit} & 18.65 & 80.4 & 227M \\
VQGAN \cite{esser2021taming} & 15.78 & 78.3 & 1.4B \\
Improved DDPM \cite{nichol2021improved} & 12.26 & - & 280M \\
ADM \cite{dhariwal2021diffusion} & 10.94 & 101.0  & 554M \\
LDM \cite{rombach2022high} & 10.56 & 103.5  & 400M \\
BigGAN-deep \cite{brock2018large} & 6.95 & 198.2  & 160M \\
MaskGIT \cite{chang2022maskgit} & 6.18 & 182.1 & 227M \\
\rowcolor{LightCyan}
\name~(ViT-B) & 6.93 & 195.8 & \grey{117M}+113M \\
\bottomrule
\end{tabular}
}
\end{center}
\vspace{-15pt}
\end{table}

\textbf{Class-Conditional Image Generation.}
Our model can also be used for class-conditional image generation as downstream task. To enable class-conditional generation, we take the ViT encoder from pre-training, and replace the original ViT decoder with a class-conditional decoder (12-layer ViT with embedding dimension 768, 113M parameters) which takes the class label as another input (concatenated to the padded features). During training, we freeze the encoder parameters to better evaluate the quality of the learned representations. Similar to pre-training, the model will take masked tokenized images as input and try to reconstruct the masked tokens. The only difference is that the decoder will not only see representations from the encoder, but also know the class label of the input image. Then during inference, the class label will be used to guide the model to generate images of the same class.

As shown in \autoref{tab:cls-cond}, \name~achieves comparable performance as SOTA image generation methods on the task of class-conditional image generation on ImageNet-1K. Note that our encoder is inherited from the pre-training and \emph{is not} fine-tuned during the downstream class-conditional training. Only the decoder is trained and has information about the class label. This shows that \name's encoder can learn high-quality representations that can achieve similar generative performance as models trained end-to-end on class conditional generative tasks.

\begin{figure*}[t]
\begin{center}
\vspace{-10pt}
\includegraphics[width=1\linewidth]{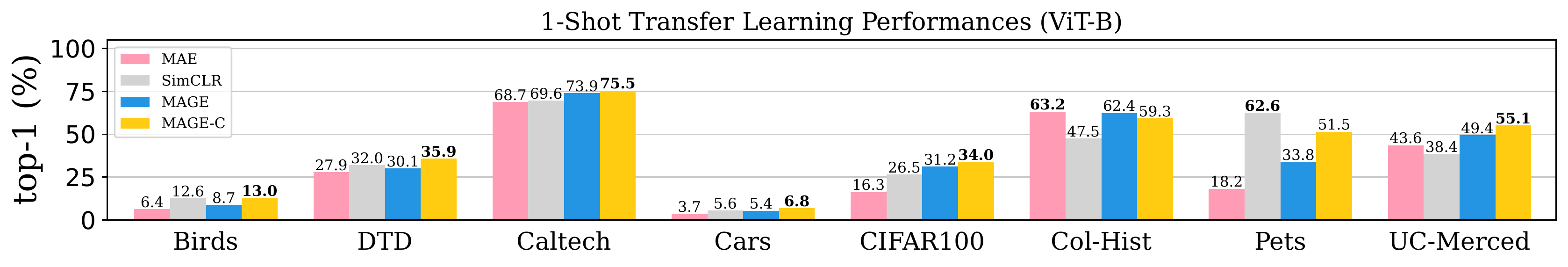}
\includegraphics[width=1\linewidth]{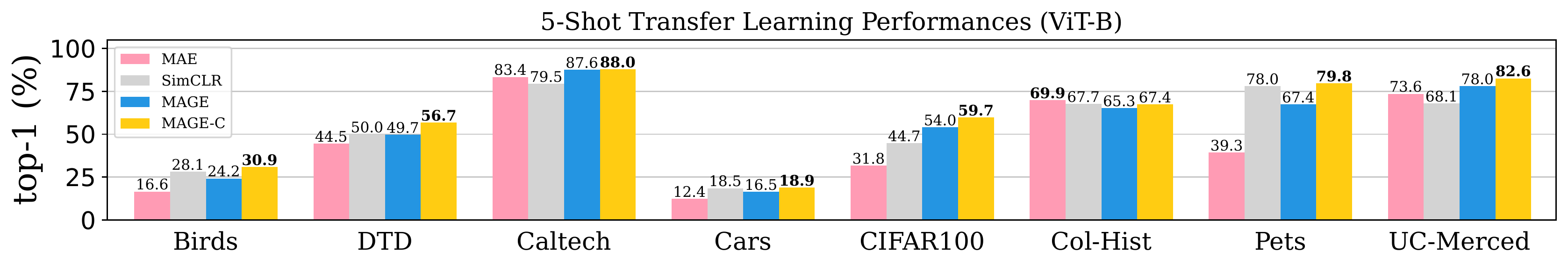}
\includegraphics[width=1\linewidth]{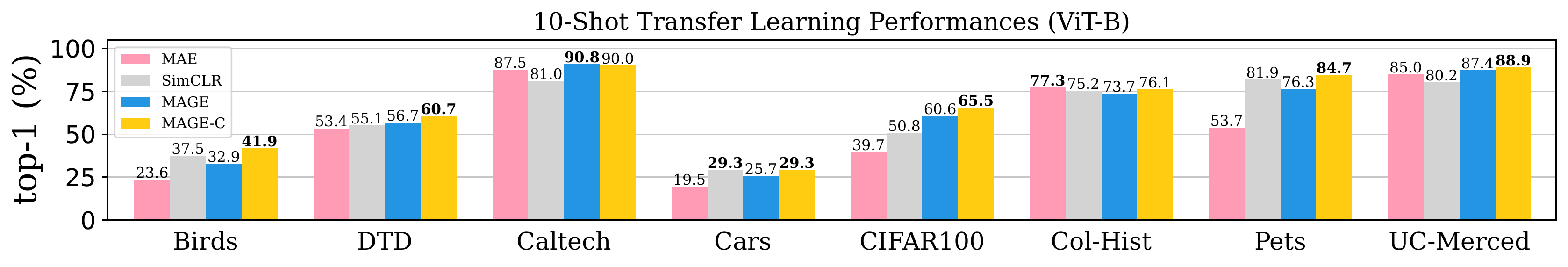}
\includegraphics[width=1\linewidth]{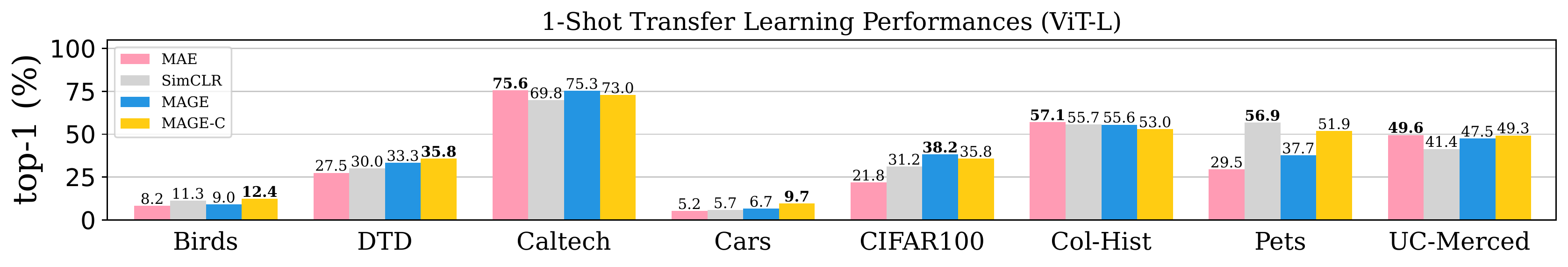}
\includegraphics[width=1\linewidth]{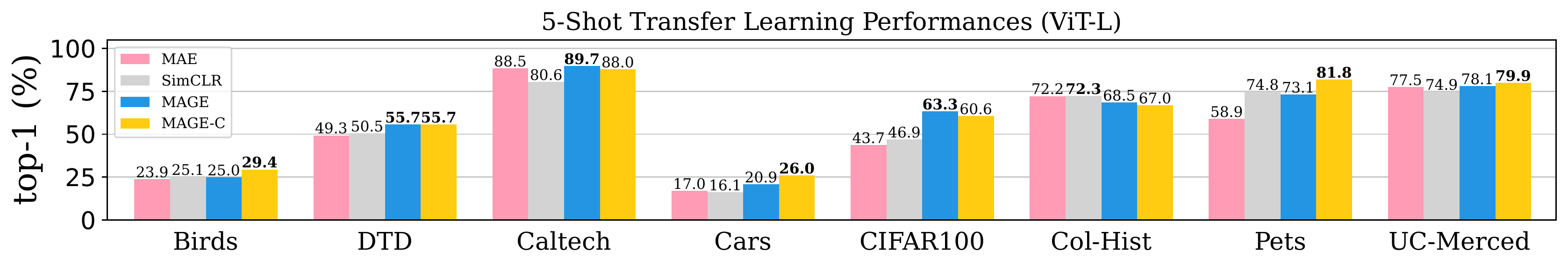}
\includegraphics[width=1\linewidth]{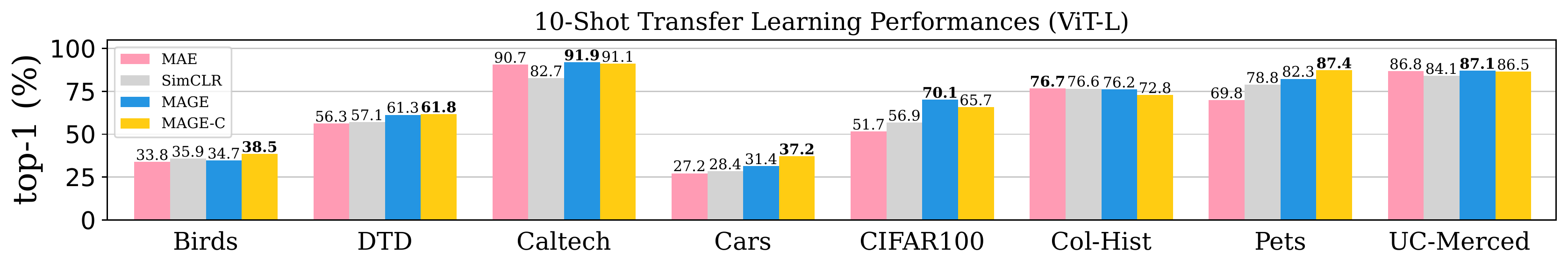}

\vspace{-18pt}
\end{center}
\caption{Transfer learning performance of ViT-B and ViT-L pre-trained on ImageNet-1K using different methods. Our methods outperform SimCLR \cite{simclr} and MAE \cite{MAE} on most datasets under different few-shot settings.}\label{fig:transfer-supp}
\vspace{-15pt}
\end{figure*}

\textbf{Few-shot Transfer Learning.} In the main paper, we provide transfer learning results of \name~on 8 different datasets with 25 samples per class. Here we further show our method's performance with 1, 5, and 10 samples per class. As shown in \autoref{fig:transfer-supp}, our method is consistently better than MAE and SimCLR on most datasets with different numbers of samples per class, demonstrating the effectiveness of our method on few-shot transfer learning.

\section{Ablation Studies}
In this section, we conduct extensive ablation studies on our method. Without further notice, we use ViT-B trained with 800 epochs for all ablation studies.

\begin{table}[h]
\caption{Top-1 accuracy of linear probing on ImageNet-1k with different method. MAE with GAN loss significantly reduces its performance on linear probing.}
\vspace{-10pt}
\label{tab:gan}
\begin{center}{
\resizebox{0.48\textwidth}{!}{
\begin{tabular}{l|cc}
\toprule
Methods & Linear Probing (\%) \\

\midrule
MAE (ViT-L) \cite{MAE} & 75.1 \\
MAE (ViT-L) + norm pixel loss \cite{MAE} & 75.8 \\
MAE + GAN loss (ViT-L) \cite{maegithub} & 64.1 \\
\name & \textbf{78.9} \\
\bottomrule
\end{tabular}
}
}
\end{center}
\vspace{-10pt}
\end{table}

\textbf{MAE with GAN loss.} One trivial solution to force the previous MIM method to generate realistic images is to add a GAN loss on top of the reconstructed image. However, we show that introducing GAN loss during previous MIM pre-training could largely decrease the performance of linear probing. As shown in \autoref{tab:gan}, we evaluate the linear probing performance of a ViT-L MAE model pre-trained with an extra GAN loss released in MAE's official GitHub repo \cite{maegithub}. Although this model can reconstruct much more realistic images than the original MAE, the linear probing performance decreases by 11\% compared with the ViT-L MAE model pre-trained without the GAN loss. On the other hand, our \name~framework enables generative modeling and representation learning to help each other, achieving SOTA performances on both tasks using one single model.

\begin{table}[h]
\caption{FID and top-1 accuracy of linear probing on ImageNet-1k by padding with \texttt{[C]} or a universal \texttt{[MASK]} token.}
\vspace{-10pt}
\label{tab:pad-clstoken}
\begin{center}{
\begin{tabular}{l|cc}
\toprule
Padding Token & FID & Linear Probing (\%) \\

\midrule
\texttt{[MASK]} & 12.4 & 72.5 \\
\texttt{[C]} & 11.6 & 73.3 \\
\bottomrule
\end{tabular}
}
\end{center}
\vspace{-10pt}
\end{table}

\textbf{Pad with [CLS] token.} To pad the output of the encoder, unlike MAE which uses a learnable mask token that is shared for different inputs, we use the class token feature which is specific to each image. This design allows the decoder to take the global features extracted by the encoder as input. As shown in \autoref{tab:pad-clstoken}, this design can improve both class-unconditional generation performance and linear-probing results.

\begin{table}[h]
\caption{FID and top-1 accuracy of linear probing of ViT-B trained 1600 epochs on ImageNet-1k using strong augmentations (s.a.) and weak augmentations (w.a.).}
\vspace{-10pt}
\label{tab:augmentations}
\begin{center}{
\begin{tabular}{l|cc}
\toprule
Augmentations & FID & Linear Probing (\%) \\

\midrule
\name~+ w.a. & \textbf{8.67} & 70.5 \\
\name~+ s.a. & 11.1 & \textbf{74.7} \\
\bottomrule
\end{tabular}
}
\end{center}
\vspace{-10pt}
\end{table}

\textbf{Augmentations.} As shown in many previous works on generative modeling and representation learning \cite{MAE, simclr, robinson2021can, donahue2019large}, the augmentation used to train the model is important for both generation and representation learning performance. In our paper, we use two different sets of augmentations: default augmentations, or strong augmentations (s.a.), which consist of random crop and resize (0.2 to 1) and random flipping; weak augmentations (w.a.), which consist of random crop and resize (0.8 to 1) and random flipping. The only difference between s.a. and w.a. is the zoom-in scale of random crop and resize. As shown in \autoref{tab:augmentations}, strong augmentations favor representation learning and weak augmentation favor generation quality, which is consistent with findings in prior works \cite{MAE, donahue2019large}.

\begin{table}[h]
\caption{FID and top-1 accuracy of linear probing of ViT-B trained 400, 800, and 1600 epochs on ImageNet-1k.}
\vspace{-10pt}
\label{tab:epochs}
\begin{center}{
\begin{tabular}{l|ccc}
\toprule
\#Pre-training Epochs & FID & Linear Probing (\%) \\
\midrule
400 & 12.2 & 72.2 \\
800 & 11.6 & 73.3 \\
1600 & 11.1 & 74.7 \\
\bottomrule
\end{tabular}
}
\end{center}
\vspace{-10pt}
\end{table}

\textbf{Pre-training Epochs.} One important factor in self-supervised learning methods is the number of pre-training epochs. Prior works have shown that longer pre-training epochs can largely improve the performance of self-supervised methods \cite{MAE, simclr}. We compare \name's performance on ViT-B using 400, 800 and 1600 epochs of pre-training in \autoref{tab:epochs}. We observe that \name~achieves good performances in both generation and representation learning with 400 epochs of pre-training, and can consistently benefit from longer training epochs.

\begin{table}[h]
\caption{FID and top-1 accuracy of linear probing on ImageNet-1k using different decoder architecture. $d$ denotes decoder depth (number of transformer blocs in the decoder), and $w$ denotes decoder width (feature dimension in the decoder).}
\vspace{-10pt}
\label{tab:decoder}
\begin{center}{
\begin{tabular}{l|ccc}
\toprule
Decoder Arch. & FID & Linear Probing (\%) \\

\midrule
$d=8, w=512$ & 12.4 & 72.1 \\
$d=8, w=768$ & 11.6 & 73.3 \\
$d=8, w=1024$ & 11.4 & 73.5 \\
\midrule
$d=6, w=768$ & 12.4 & 71.8 \\
$d=8, w=768$ & 11.6 & 73.3  \\
$d=10, w=768$ & 11.4 & 73.2 \\
$d=12, w=768$ & 11.3 & 73.4 \\
\bottomrule
\end{tabular}
}
\end{center}
\vspace{-10pt}
\end{table}

\textbf{Decoder Design.} MAE \cite{MAE} shows that a small ViT decoder is enough to achieve good performance. We also try different decoder architectures and summarize the results in \autoref{tab:decoder}. As shown in the table, the decoder with 8 blocks and 768 feature dimension reaches the best balance between computation cost and performance for ViT-B. Therefore, we choose the decoder architecture to be 8 blocks with 768 feature dimensions for ViT-B and 1024 feature dimension for ViT-L in the paper.

\begin{table}[h]
\caption{Top-1 accuracy of linear probing on ImageNet-1k using different methods. C denotes our contrastive loss, R denotes our reconstructive loss, $\dagger$ denotes our re-implementation.}
\vspace{-10pt}
\label{tab:contrastive}
\begin{center}{
\begin{tabular}{l|ccc}
\toprule
Methods  & Linear Probing (\%) \\
\midrule

MAE \cite{MAE} & 68.0 \\
R only & 73.3 \\
SimCLR $^\dagger$ \cite{simclr} & 74.2 \\
C only & 72.9 \\
C+R (\name-C) & \textbf{77.1} \\
\bottomrule
\end{tabular}
}
\end{center}
\vspace{-10pt}
\end{table}

\textbf{Complement MIM with Contrastive Loss.} We show in the main paper that \name~can be further combined with a simple contrastive loss (\name-C) to achieve better representation learning performance. In \autoref{tab:contrastive} we show more ablations regarding this contrastive loss. The performance of simply applying the contrastive loss without the reconstructive loss is worse than the SimCLR baseline. This is likely because we do not use augmentations such as color jittering and random grey scale, so applying only the contrastive loss could result in learning shortcut semantics such as color distribution \cite{simclr, robinson2021can}. However, the reconstructive loss can prevent the network from falling into such shortcut solutions and help the network learn richer semantics.

\begin{table}[h]
\caption{FID and top-1 accuracy of linear probing of \name-C on ImageNet-1k using different maximum masking ratio $\max(m_r)$.}
\vspace{-10pt}
\label{tab:contrastive-maxmr}
\begin{center}{
\begin{tabular}{l|ccc}
\toprule
 & FID & Linear Probing (\%) \\
\midrule
$\max(m_r)$=1.0 & 14.1 & 75.0 \\
$\max(m_r)$=0.7 & 23.5 & 76.3 \\
$\max(m_r)$=0.6 & 27.0 & 77.1 \\
\bottomrule
\end{tabular}
}
\end{center}
\vspace{-10pt}
\end{table}

We also notice one problem of applying contrastive training to \name: \name~can see very high masking ratios during training, but applying positive loss to two augmented views of the same image both with a very high masking ratio is problematic. This is because such two views can share very little common information, leading to a performance drop as shown in \cite{tian2019contrastive}. Therefore, we only apply contrastive loss when the masking ratio is relatively low ($m_r<0.6$). In \autoref{tab:contrastive-maxmr}, we show the performance of generation and representation learning w.r.t. the maximum masking ratio of our variable masking ratio distribution. Smaller $\max(m_r)$ leads to better linear probing but worse FID. We believe it is because, with smaller $\max(m_r)$, the contrastive loss can operate on more samples in the batch whose masking ratio $m_r<0.6$, which is important for contrastive learning as shown in \cite{simclr}. On the other hand, small $\max(m_r)$ harms the generation performance because the network should see a relatively high masking ratio to enable generation from blank image (100\% masking ratio). We leave a further investigation of this phenomenon and a better combination strategy for future work.

\section{Implementation Details}

\textbf{Tokenizer and Detokenizer.} We use a CNN-based VQGAN encoder and quantizer to tokenize the 256x256 input images to 16x16 discrete tokens. The detokenizer operates on the 16x16 discrete tokens and reconstructs the 256x256 image. The encoder consists of 5 blocks and each block consists of 2 residual blocks. After each block in the encoder, the feature is down-sampled by 2 using average pooling. The quantizer then quantizes each pixel of the encoder's output feature map using a codebook with 1024 entries, each entry with dimension 256. The detokenizer consists of another 5 blocks where each block consists of 2 residual blocks. After each block in the decoder, the feature map is up-sampled by 2. Please refer to our code and the original VQGAN paper for more details \cite{esser2021taming}.

\textbf{ViT architecture.} After the tokenizer, the latent sequence length becomes 256 (plus one 'fake' class token). We then follow a similar encoder-decoder Transformer architecture similar to MAE \cite{MAE}. More specifically, we use standard ViT architecture \cite{vit}, which consists of a stack of Transformer blocks \cite{vaswani2017attention}, where each block consists of a multi-head self-attention block and an MLP block. We use two learnable positional embeddings, one added to the input of the encoder and another added to the input of the decoder. 

We use features from the encoder output for classification tasks, such as linear probing, few-shot transfer learning, and fine-tuning. We average pool the encoder output without the class token to get the input of the linear classifier.

\begin{table}[t]
\caption{\textbf{Pre-training Setting.}}
\vspace{-10pt}
\label{tab:pretrain-setting}
\begin{center}{
\small
\begin{tabular}{l|l}

config & value \\
\toprule

optimizer & AdamW \cite{loshchilov2017decoupled} \\
base learning rate & 1.5e-4 \\ 
weight decay & 0.05 \\
optimizer momentum & $\beta_1, \beta_2=0.9, 0.95$ \\
batch size & 4096 (B), 2048 (L) \\
learning rate schedule & cosine decay \cite{loshchilov2016sgdr} \\
warmup epochs & 40 \\
training epochs & 1600 \\
gradient clip & 3.0 \\
label smoothing \cite{szegedy2016rethinking} & 0.1 \\
dropout & 0.1 \\
masking ratio min & 0.5 \\
masking ratio max & 1.0 (\name) 0.6 (\name-C) \\
masking ratio mode & 0.55 \\
masking ratio std & 0.25 \\
\midrule
\textit{ \name-C only} \\
contrastive loss weight & 0.1 \\
temperature & 0.2 \\

\end{tabular}
}
\end{center}
\vspace{-10pt}
\end{table}

\begin{table}[t]
\caption{\textbf{Linear Probing Setting.}}
\vspace{-10pt}
\label{tab:lp-setting}
\begin{center}{
\small
\begin{tabular}{l|l}

config & value \\
\toprule

optimizer & LARS \cite{you2017large} \\
base learning rate & 0.1 (B) 0.05 (L) \\ 
weight decay & 0 \\
optimizer momentum & 0.9 \\
batch size & 4096 \\
learning rate schedule & cosine decay \cite{loshchilov2016sgdr} \\
warmup epochs & 10 \\
training epochs & 90 \\
augmentation & RandomResizedCrop \\

\end{tabular}
}
\end{center}
\vspace{-10pt}
\end{table}

\begin{table}[t]
\caption{\textbf{End-to-end fine-tuning Setting.}}
\vspace{-10pt}
\label{tab:finetune-setting}
\begin{center}{
\small
\begin{tabular}{l|l}

config & value \\
\toprule

optimizer & AdamW \cite{loshchilov2017decoupled} \\
base learning rate & 2.5e-4 \\ 
weight decay & 0.05 \\
optimizer momentum & $\beta_1, \beta_2=0.9, 0.999$ \\
layer-wise lr decay \cite{bao2021beit} & 0.65 (B) 0.75 (L) \\
batch size & 1024 \\
learning rate schedule & cosine decay \cite{loshchilov2016sgdr} \\
warmup epochs & 5 \\
training epochs & 100 (B) 50 (L) \\
label smoothing \cite{szegedy2016rethinking} & 0.1 \\
augmentation & RandAug (9, 0.5) \cite{cubuk2020randaugment} \\
mixup \cite{zhang2017mixup} & 0.8 \\
cutmix \cite{yun2019cutmix} & 1.0 \\
random erase & 0 \\
drop path \cite{huang2016deep} & 0.1 (B) 0.2 (L)

\end{tabular}
}
\end{center}
\vspace{-10pt}
\end{table}

\begin{table}[t]
\caption{\textbf{Supervised training from scratch setting with ViT on semantic tokens.}}
\vspace{-10pt}
\label{tab:scratch-setting}
\begin{center}{
\small
\begin{tabular}{l|l}

config & value \\
\toprule

optimizer & AdamW \cite{loshchilov2017decoupled} \\
base learning rate & 1e-4 \\ 
weight decay & 0.3 \\
optimizer momentum & $\beta_1, \beta_2=0.9, 0.95$ \\
batch size & 4096 \\
learning rate schedule & cosine decay \cite{loshchilov2016sgdr} \\
warmup epochs & 20 \\
training epochs & 300 (B) 200 (L) \\
label smoothing \cite{szegedy2016rethinking} & 0.1 \\
augmentation & RandAug (9, 0.5) \cite{cubuk2020randaugment} \\
mixup \cite{zhang2017mixup} & 0.8 \\
cutmix \cite{yun2019cutmix} & 1.0 \\
drop path \cite{huang2016deep} & 0.1 (B) 0.2 (L) \\
exp. moving average (EMA) & 0.9999

\end{tabular}
}
\end{center}
\vspace{-10pt}
\end{table}

\textbf{Pre-training.} Please refer to \autoref{tab:pretrain-setting} for our default pre-training setting. We use only random crop and resize (0.2 to 1) and random horizontal flip as our default augmentations.

\textbf{Generation.} We use iterative decoding as in MaskGIT \cite{chang2022maskgit} to iteratively fill in masked tokens and generate images.  To generate an image at inference time, we start from a blank canvas with all the tokens masked out, i.e., $Y_M^{(0)}$. For iteration $t=1, \cdots, T$, the algorithm runs as follows:

1. \textbf{Predict.} Given $Y_M^{(t)}$ which is the unmasked tokens at the beginning of iteration $t$, we first predict the probability of the remaining masked tokens using our model, denoted as $p^{(t)}\in \mathbb{R}^{N_t\times K}$, where $N_t$ is the number of remaining masked tokens and $K$ is the number of entries in the codebook.

2. \textbf{Sample.} At each masked location $i$, we sample a token $y_i^{(t)}$ based on the prediction probability $p_i^{(t)}\in \mathbb{R}^{K}$ over all possible tokens in the codebook, and form the unmasked prediction $Y^{(t)}$. After $y_i^{(t)}$ is sampled, its corresponding prediction score plus a noise sampled from a random Gumbel distribution multiplied by temperature $\tau$ is used as the "confidence" score indicating the model's belief of the prediction at location $i$. The confidence scores at the unmasked locations are set to $+\infty$.

3. \textbf{Mask.} We then determine the number of tokens $N_{t+1}$ for the next iteration $t+1$ based on a cosine masking schedule $N_{t+1} = N_0 \cdot \cos(\frac{\pi t}{2T})$. We then sample $N_{t+1}$ locations with the lowest confidence scores and mask those locations from $Y^{(t)}$ to generate $Y_M^{(t+1)}$.

For class unconditional generation, we choose $\tau=6.0$ and $T=20$ to generate images in our experiment.

\textbf{Linear Probing \& Fine-tuning.} Our linear probing and fine-tuning setup follow MAE \cite{MAE}. Please see \autoref{tab:lp-setting} and \autoref{tab:finetune-setting} for detailed configurations.

\textbf{Training from scratch.} We also follows MAE \cite{MAE} for our training-from-scratch baseline. \autoref{tab:scratch-setting} summarizes our configurations.

\textbf{Code.} For more implementation details, please refer to our code at \href{https://github.com/LTH14/mage}{\textcolor{red}{\texttt{https://github.com/LTH14/mage}}}.


\end{document}